%% file: 0-main.tex
\documentclass{article}

 \usepackage[preprint]{neurips_2026}

\usepackage[utf8]{inputenc} % allow utf-8 input
\usepackage[T1]{fontenc}    % use 8-bit T1 fonts
\usepackage{hyperref}       % hyperlinks
\usepackage{url}            % simple URL typesetting
\usepackage{booktabs}       % professional-quality tables
\usepackage{amsfonts}       % blackboard math symbols
\usepackage{nicefrac}       % compact symbols for 1/2, etc.
\usepackage{microtype}      % microtypography
\usepackage{xcolor}         % colors
\usepackage{multirow}
\usepackage{graphicx}
\usepackage{amsmath} 
\usepackage{float}
\usepackage{array}
\usepackage[table]{xcolor}
\usepackage{colortbl}
\usepackage{makecell}

\title{$\mathsf{BenchHAR}$: Benchmarking Self-Supervised Learning for Generalizable Sensor-based Activity Recognition}

\author{%  
    Yize Cai$^1$, 
    Rui Feng$^1$, 
    Anlan Yu$^2$, 
    Baoshen Guo$^3$, 
    Zhiqing Hong$^1$ \\
 $^1$The Hong Kong University of Science and Technology (Guangzhou), $^2$Peking University \\
 $^3$Singapore-MIT Alliance for Research and Technology\\
  \texttt{ycai186@connect.hkust-gz.edu.cn, fengr382@gmail.com} \\
  \texttt{yal6040@pku.edu.cn, baoshen@mit.edu, zhiqinghong@hkust-gz.edu.cn} \\
}

\begin{document}

\maketitle

\begin{abstract}
Human Activity Recognition (HAR) from wearable sensors supports broad healthcare and behavior science applications.
However, data heterogeneity and the scarcity of labeled data limit its real-world generalization.
Recent advances in self-supervised learning (SSL) in vision and language domains have shown strong capability for learning generalizable representations from unlabeled data.
Yet, few principled studies have systematically compared the generalization performance of SSL methods or explored how to adapt them for generalizable HAR.
To address these gaps, we present $\mathsf{BenchHAR}$, a unified framework for evaluating the generalization capability of SSL methods for sensor-based HAR on unseen target distributions.
$\mathsf{BenchHAR}$ curates a large-scale dataset ($\sim$258K samples) and evaluates eight representative SSL methods across 12 encoder-classifier architectures.
Our results reveal that existing SSL methods still struggle to achieve satisfactory generalization performance.
We find that: 
(1) For HAR models, the hybrid paradigm (combining reconstruction and contrastive pretraining) achieves the best overall performance.
The CNN encoder exhibits the strongest ability to learn generalizable sensor representations, while more expressive classifier architectures further improve generalization.
(2) For data scale, increasing the amount of pretraining data from downstream activity classes consistently improves generalization, while adding more labeled data yields only limited gains.
Interestingly, incorporating unlabeled data from non-downstream activity classes does not improve generalization.
(3) Data collected from custom-grade devices generalizes better than that from research-grade devices, and sensor data from the limb transfers more effectively to trunk positions. 
Overall, $\mathsf{BenchHAR}$ provides a unified benchmark and actionable insights for developing generalizable sensor-based HAR systems.
% Our code is available at \href{https://anonymous.4open.science/r/HAR-Bench-B9D7/README.md}{\textcolor{blue}{\textit{this link}}}.
Our code is available at \href{https://github.com/saiketa/HAR-Bench}{\textcolor{blue}{\textit{this link}}}.
\end{abstract}

\input{1-intro}

\input{2-relatedwork}

\input{3-methods}

\input{4-experiments}

\input{5-discussion}

\input{6-conclusion}

\bibliographystyle{plain}
\bibliography{ref}

% \newpage
% \input{checklist.tex}

\input{7-appendix}

\end{document}

%% file: 1-intro.tex
\section{Introduction}

With the rapid proliferation of sensor-embedded wearable devices such as smart rings, smartwatches, and smartphones, Human Activity Recognition (HAR) based on Inertial Measurement Units (IMUs) has attracted increasing attention and has emerged as a solid foundation for human-centered computing.
Sensor-based HAR enables a wide spectrum of practical applications.
In healthcare, it enables critical functions in a continuous and cost-effective manner, such as health monitoring~\cite{moreau2023overview, straczkiewicz2021systematic} and risk evaluation~\cite{biswas2025wearable, khurshid2022wearable}. 
In embodied intelligence, it provides a critical perceptual layer that allows machines to understand and respond to human behaviors~\cite{grandi2025virtual, chen2026noise}.
On the societal scale, large-scale activity information offers a unique lens into human life patterns, informing urban planning and revealing collective behavioral dynamics~\cite{xu2025experience, hong2025experience, althoff2025countrywide, gao2025wearable}. 
These important applications show strong demand for designing accurate and robust sensor-based HAR models, making \textbf{Generalizable HAR} an increasingly critical research focus in recent years~\cite{hong2025llm4har, hong2024crosshar, yan2025large, lu2022semantic, cai2025towards}.

\begin{figure}[t]
	\centering
	\includegraphics[width=5.5in]{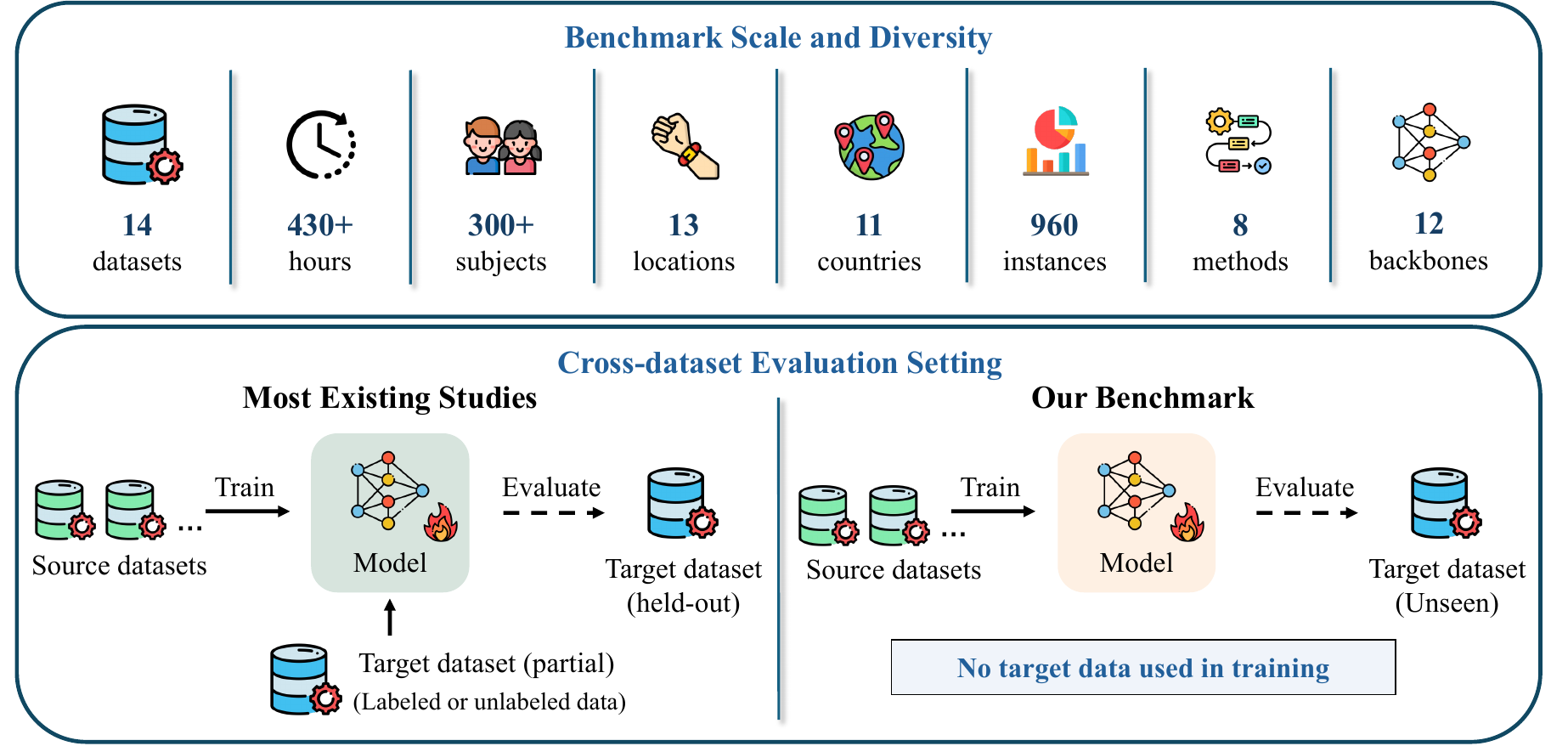}
	\caption{Benchmark overview. 
    % Cross-dataset evaluation means training with source datasets and testing with unseen target datasets. This setting follows the practical activity recognition scenario. 
    Unlike most existing cross-dataset studies that utilize target datasets during training, $\mathsf{HAR\text{-}Bench}$ evaluates generalization by training on source datasets and testing on unseen target datasets, reflecting realistic sensor-based HAR deployment scenarios.} 
	\label{fig:framework}
\end{figure}

However, developing generalizable sensor-based HAR models remains fundamentally constrained by two key challenges.
\textit{(i) Cross-domain data heterogeneity.}
Due to the embodied nature of wearable devices, sensor data is subject to substantial heterogeneity across domains, such as different user habits, device types, and sensor placements~\cite{cai2025towards, xu2021limu}.
Such heterogeneity introduces significant domain discrepancies between the training and testing data, which commonly arise in real-world deployments and lead to cross-dataset generalization problem~\cite{zhang2025mopformer, hong2024crosshar, hong2025llm4har, yan2025large}.
\textit{(ii) Sensor data scarcity.}
Unlike vision and language domains with large amounts of readily accessible data, existing HAR datasets are typically small in scale and limited in diversity. 
Moreover, the non-intuitive nature of sensor data leads to high annotation costs.
Such datasets are insufficient to capture the complexity and variability of real-world human activities, limiting the learning of robust and generalizable representations.

In recent years, the success of self-supervised learning (SSL) methods in vision~\cite{chen2020simple} and language~\cite{devlin2019bert} domains for learning generalizable representations from large-scale unlabeled data has inspired their application to sensor-based HAR.
However, unlike discrete tokens or pixel grids, sensor data consists of continuous physical measurements with strong multivariate and multimodal coupling. 
It remains unclear how SSL methods perform under such unique data and challenging generalization scenarios.
This calls for a comprehensive benchmark of key design choices, such as SSL paradigms, methods, architectures, and data recipes, to systematically evaluate their effectiveness.
This study seeks to fill these gaps and improve generalization in sensor-based HAR by exploring two key directions:

\textbf{(i) Integrating multiple datasets for large-scale training.} 
Recent studies have explored mitigating data scarcity by integrating multiple datasets, such as in physiological data~\cite{singh2025feel} and respiratory audio~\cite{zhang2024towards}. 
Over the past decade, the HAR community has developed a wide range of publicly available datasets~\cite{haresamudram2025past}. 
These datasets are typically small in scale and highly diverse in sensor configurations and collection protocols, providing a strong foundation for dataset integration. 
Despite this potential, only a limited number of studies~\cite{napoli2024benchmark, zhang2025mopformer} explore dataset integration on a limited scale for HAR, and large-scale dataset curation and joint training remain largely underexplored. 

\textbf{(ii) SSL for generalizable HAR.} 
The ability of SSL to learn generalizable representations makes it a promising solution for improving generalizable HAR.
More importantly, learning from unlabeled data holds promise for reducing the reliance on data annotations, which is particularly beneficial for sensor data due to the high cost and inherent ambiguity of labeling. 
However, most existing SSL-based HAR studies rely on isolated evaluation protocols~\cite{xu2021limu, yuan2024self, hong2024crosshar}, making fair and systematic comparisons difficult. 
Although SSL benchmarks for sensor-based HAR have been proposed~\cite{haresamudram2022assessing, qian2022makes}, they typically assume access to target datasets during training (Figure~\ref{fig:framework} left), and thus fail to evaluate the capability of SSL under generalization scenarios.

In this paper, we propose $\mathsf{BenchHAR}$, a novel benchmark for evaluating the generalization capability of SSL sensor-based HAR methods. 
As shown in Figure~\ref{fig:framework}, $\mathsf{BenchHAR}$ aggregates diverse and heterogeneous HAR datasets to enable large-scale joint pretraining, and systematically evaluates state-of-the-art (SOTA) SSL methods on unseen datasets to assess their generalization capability.
Our contributions are summarized as follows:

\begin{itemize}
    \item We curate and standardize a large-scale ($\sim$258K samples, $\sim$430 hours), multi-source (14 widely used datasets), diverse (300+ subjects, 11 countries, 13 locations and 2 device types), and multimodal (accelerometer and gyroscope) sensor-based HAR dataset, which exhibits substantially greater diversity than existing single-source or small-scale curated datasets.
    \item We investigate four pretraining paradigms (recognition, reconstruction, contrastive, and hybrid) to evaluate the cross-dataset generalization capability of SSL HAR methods.
    The evaluation involves eight SSL methods, 12 widely used model backbones, and two modality combinations, enabling a systematic analysis of the impact of training design choices.
    \item We systematically investigate how data scale (both unlabeled and labeled data) and domain shift factors (cross-subject, cross-location, and cross-device) affect model performance, providing insights for future sensor-based HAR research.
\end{itemize}

Through extensive experiments, we obtain the following \textbf{key findings}. 
These findings may offer useful insights for future research on generalizable HAR systems and sensor foundation models.
\begin{itemize}
    \item (1)  Unlike in-dataset HAR studies that achieve over 80\% F1 performance, current SSL methods struggle to achieve satisfactory cross-dataset generalization performance, with the best F1 reaching 61.31\%. 
    This highlights the strong need for further research in this area.
    \item (2) For HAR models, the hybrid paradigm (combining reconstruction and contrastive pretraining) achieves the best overall performance.
    CNN encoders exhibit strong capability in learning generalizable representations, while more expressive classifiers (e.g., CNNs, GRUs, and Transformers) further improve generalization performance.
    \item (3) For data scale, increasing the pretraining data size from downstream activity classes consistently improves generalization, while adding more labeled data for downstream training yields limited gains. 
    However, incorporating unlabeled data from non-downstream activity classes for pretraining does not improve generalization.
    \item (4) Sensor data collected from consumer-grade devices generalizes better to research-grade devices, and data from limb locations transfers more effectively to trunk positions. 
    These findings provide useful insights for future sensor data collection.
\end{itemize}

%% file: 2-relatedwork.tex
\section{Related Work}

\subsection{Sensor-based Human Activity Recognition}

Sensor-based Human Activity Recognition (HAR) aims to classify daily human activities from wearable sensor data. 
To enable robust and generalizable deployment of HAR systems, existing studies have explored addressing \textit{cross-domain data heterogeneity} and \textit{sensor data scarcity}.
Early studies primarily focus on the single domain shift factor, such as cross-subject~\cite{qin2023generalizable}, cross-location~\cite{lu2024diversify}, and cross-device~\cite{ahmad2024hyperhar} within the same dataset. 
However, for real-world deployment, these factors typically co-occur, resulting in more complex scenarios. %that are not adequately captured by single cross-domain factor settings.
Recent efforts have begun to explore cross-dataset HAR, a more realistic setting where models are trained and evaluated on different datasets. 
Some methods try to address cross-dataset domain shifts by accessing target datasets during training~\cite{hu2023swl, xu2023practically, wang2024optimization, xue2025mobhar}. 
However, such assumptions are often impractical in real-world deployments, where target domain data is often unavailable~\cite{hong2024crosshar}. 
Other methods attempt to generalize to unseen target datasets using only source datasets to learn generalizable representations~\cite{hong2024crosshar, lu2022semantic, xiong2025generalizable}. 
Nevertheless, these methods are typically limited to small-scale datasets, making them prone to overfitting to source dataset distributions and resulting in suboptimal performance.
To address these limitations, we propose a comprehensive multi-dataset training framework that jointly leverages existing small-scale datasets to enable a more realistic evaluation of cross-dataset generalization.

\subsection{Sensor Data Pretraining}

Self-supervised learning (SSL) has demonstrated strong potential for learning generalizable representations from unlabeled data~\cite{girdhar2023imagebind, devlin2019bert}. 
Given the high cost of annotating sensor data, SSL has emerged as a promising direction to improve the generalization capability of sensor-based HAR models~\cite{logacjov2024self}. 
Recent studies have explored various SSL paradigms for sensor-based HAR, such as masked reconstruction~\cite{xu2021limu, zhang2025mopformer, xu2025experience}, contrastive learning~\cite{qian2022makes, liu2023focal, zhang2024unimts, dai2024contrastsense}, and other objectives~\cite{hong2024crosshar, yuan2024self}. In addition, several general time-series pretraining methods have also shown competitive performance on sensor data~\cite{ijcai2021-324, zhang2023self, yue2022ts2vec, dong2023simmtm}.
Despite these advances, the use of self-formulated evaluation protocols and the assumption of access to target dataset during training prevent existing studies from fairly and effectively evaluating the generalization performance of SSL methods.
To address these gaps, this study establishes a comprehensive benchmark to systematically and fairly evaluate SSL methods for cross-dataset generalization without access to target datasets during training.

\subsection{Benchmarks in Sensor-based HAR}

\input{tables/benchmarks}

A summary of sensor-based HAR benchmarks is shown in Table~\ref{tab:benchmark}.
Traditional benchmarks for sensor-based HAR primarily train and evaluate models within individual datasets~\cite{hossain2025benchmarking, abdel2021human}, which fails to reflect real-world scenarios. 
For SSL methods, CL-HAR~\cite{qian2022makes} benchmarks contrastive learning methods within four individual datasets.
Haresamudram et al.~\cite{haresamudram2022assessing} and Ek et al.~\cite{ek2025comparing} evaluate SSL methods in settings where target datasets are accessible during training.
Recently, DAGHAR~\cite{napoli2024benchmark} and HAROOD~\cite{lu2026harood} benchmark representation learning methods for cross-dataset generalization without training on target datasets, but do not involve SSL methods.
Therefore, few benchmarks comprehensively evaluate the cross-dataset generalization capability of SSL HAR methods.
In addition, most existing sensor-based HAR benchmarks are generally limited in data scale, restricting models from learning generalizable sensor representations.
Although Haresamudram et al.~\cite{haresamudram2022assessing} employ a large-scale sensor dataset~\cite{chan2024capture}, it still lacks sufficient diversity in subjects, device types, and sensor placements.
In contrast to existing benchmarks, we construct a large-scale dataset that contains diverse activity classes, subjects, device types, and sensor placements, to enable comprehensive evaluations of a wide range of SSL methods under the cross-dataset generalization setting.

%% file: tables/benchmarks.tex
\begin{table}[t]
\centering
\footnotesize
\caption{Comparison of sensor-based HAR benchmarks.}
\scalebox{0.9}{
\begin{tabular}{c|cc|cccc|cccc}
\toprule
\multirow{2}{*}{Benchmark} 
& \multicolumn{2}{c|}{Data} 
& \multicolumn{4}{c|}{Self-supervised} 
& \multicolumn{4}{c}{Cross-domain factor} \\ 
\cmidrule(lr){2-3} \cmidrule(lr){4-7} \cmidrule(l){8-11}
& \# Dataset & Joint training 
& Rec. & Recon. & Contr. & Hybrid 
& User & Position & Device & Dataset \\ 
\midrule
~\cite{abdel2021human} & 6 & $\times$ &  &  &  &  & $\checkmark$ &  &  & \\
DAGHAR~\cite{napoli2024benchmark} & 6 & Small-scale &  &  &  &  & $\checkmark$ &  &  & $\checkmark$ \\
\cite{hossain2025benchmarking} & 5 & $\times$ &  &  &  &  &  &  &  & \\
HAROOD~\cite{lu2026harood} & 6 & $\times$ &  &  &  &  & $\checkmark$ & $\checkmark$ &  & $\checkmark$ \\ 
\midrule
\cite{haresamudram2022assessing} & 10 & $\times$ & $\checkmark$ & $\checkmark$ & $\checkmark$ &  & $\checkmark$ & $\checkmark$ &  & \\
CL-HAR~\cite{qian2022makes} & 3 & $\times$ &  &  & $\checkmark$ &  & $\checkmark$ & $\checkmark$ &  & \\
\cite{ek2025comparing} & 6 & Small-scale &  & $\checkmark$ & $\checkmark$ &  & $\checkmark$ &  &  & \\ 
\midrule
\textbf{Ours} & 14 & Large-scale & $\checkmark$ & $\checkmark$ & $\checkmark$ & $\checkmark$ & $\checkmark$ & $\checkmark$ & $\checkmark$ & $\checkmark$ \\ 
\bottomrule
\end{tabular}
}
\label{tab:benchmark}
\end{table}

%% file: 3-methods.tex
\section{Benchmark Design}
\label{sec:design}

\subsection{Benchmarking Datasets}

\textbf{Data Curation.}
To alleviate data scarcity in sensor-based HAR, we curate multiple heterogeneous and publicly available datasets into a comprehensive benchmark.
Specifically, we investigate 25 publicly available HAR datasets and select 14 datasets based on two key criteria:
(i) Each dataset provides accelerometer and gyroscope data, which are essential for capturing human motion dynamics~\cite{chen2021deep}.
(ii) These datasets are widely adopted in the existing HAR literature~\cite{zhang2025mopformer, park2024calanet, xu2021limu, zhang2024unimts}, ensuring their reliability and representativeness.
These datasets exhibit substantial diversity across demographic populations, collection protocols, and activity classes, providing a realistic and comprehensive testbed for evaluating generalization capability of HAR models.
Detailed dataset investigations and summaries are provided in Appendix~\ref{app:overview}.

\textbf{Standardization and Preprocessing.}
Due to substantial variations in collection protocols and annotation schemes across datasets, we perform a unified standardization process over all 14 datasets.
We normalize sensor placement by mapping wearable devices to a set of consistent body-location categories under a unified naming convention.
For activity labels, we identify significant discrepancies in label definitions across datasets and consolidate and harmonize them into a unified taxonomy to ensure consistency across datasets.

Following common practice in existing studies~\cite{hong2024crosshar, hong2025llm4har, xu2021limu}, we segment continuous sensor streams into non-overlapping windows of 6 seconds. 
All data are resampled to 20 Hz, providing sufficient temporal resolution for HAR while remaining efficient for deployment on mobile and wearable devices~\cite{xu2021limu}. 
We apply instance normalization~\cite{zhou2023one, kim2021reversible} to each sample, preserving the relative data shape to mitigate the absolute magnitude variations caused by different device configurations and behavioral habits.
Each sample is annotated with a multi-dimensional label vector, including activity ID, subject ID, body-location ID, and dataset ID.

\begin{figure}[t]
	\centering
	\includegraphics[width=5.6in]{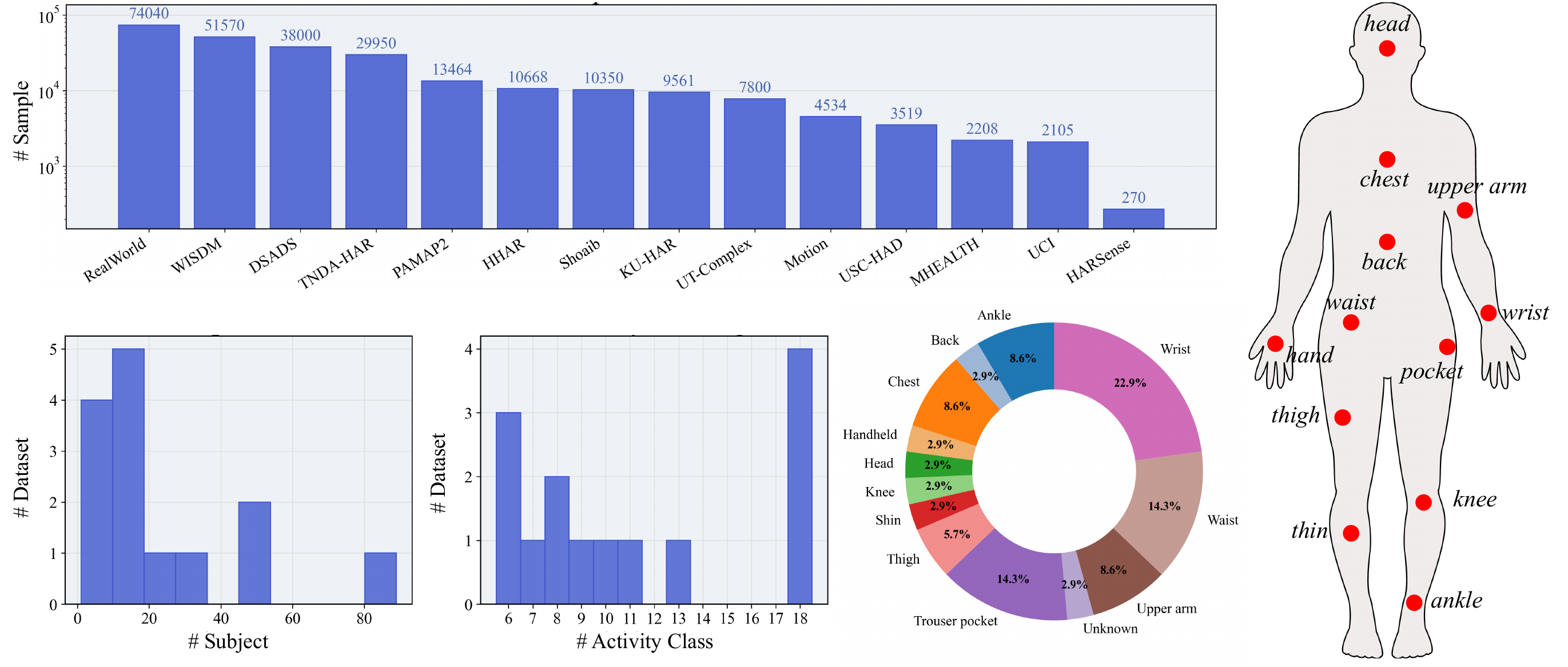}
	\caption{Benchmark dataset statistics.
    (top) Sample size distribution of each dataset.
    (bottom left) Subject count distribution.
    (bottom middle) Activity class distribution.
    (bottom right) Sample size distribution of each body location.
    (right) Illustration of body locations.} 
	\label{fig:dataset}
\end{figure}

In total, the resulting benchmark comprises over 258K samples ($\sim$430 hours) of sensor data from over 300 subjects across 11 countries, covering 13 body locations and 62 activity classes.
Figure~\ref{fig:dataset} and Appendix~\ref{app:statistics} summarize the statistics of the curated datasets.

\subsection{Pretraining Paradigms}

SSL has demonstrated strong capability in learning generalizable representations from large-scale unlabeled data across diverse domains. 
In this work, we aim to systematically evaluate the effectiveness of advanced SSL methods for cross-dataset generalization in sensor-based HAR.
Specifically, we consider four representative SSL paradigms that have been extensively studied in both sensor and general time-series tasks and have shown strong generalization performance. 

\begin{figure}[t]
	\centering
	\includegraphics[width=5.6in]{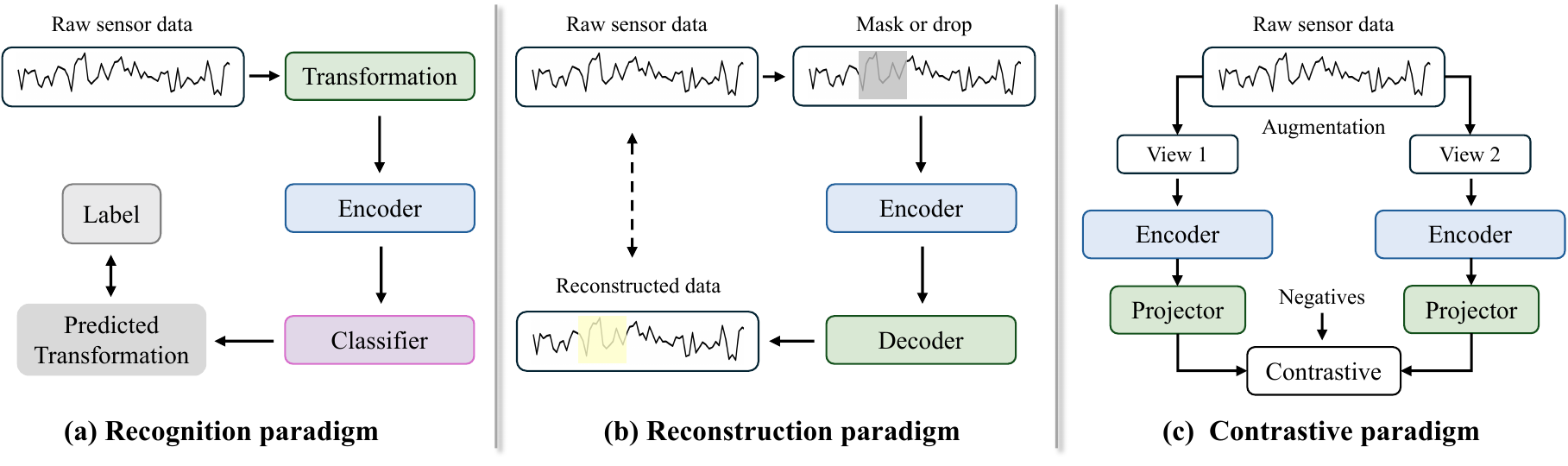}
	\caption{Pretraining paradigms for sensor-based HAR.} 
	\label{fig:ssl}
\end{figure}

\textbf{Recognition Paradigm.}
As shown in Figure~\ref{fig:ssl} (a), recognition paradigm trains the encoder using transformation recognition objectives, where the model is tasked with identifying the types of data transformation applied to input samples.
% This paradigm aims to capture semantic information in sensor data by distinguishing variations induced by such transformations~\cite{saeed2019multi}.
We adopt BioBankSSL~\cite{yuan2024self}, a SOTA framework of the recognition paradigm, as a representative method.

\textbf{Reconstruction Paradigm.}
As shown in Figure~\ref{fig:ssl} (b), reconstruction paradigm learns generalizable representations by using input reconstruction as the pretraining objective, where the model is trained to recover cropped or masked portions of the input data.
% This formulation encourages the model to capture local temporal dependencies and intrinsic structure in sensor data.
We select LIMU-BERT~\cite{xu2021limu}, a BERT-style~\cite{devlin2019bert} masked reconstruction method tailored for sensor data, and CRT~\cite{zhang2023self}, a time-frequency dropped reconstruction method designed for general time-series analysis, as representative methods.

\textbf{Contrastive Paradigm.}
As shown in Figure~\ref{fig:ssl} (c), contrastive paradigm learns representations by constructing positive and negative pairs, encouraging the model to minimize the distance between positive samples while maximizing the distance from negative samples in the representation space. 
% This objective enables the learning of discriminative and globally consistent representations.
We select TS-TCC~\cite{ijcai2021-324} (a context-oriented contrastive method), TS2Vec~\cite{yue2022ts2vec} (a hierarchical contrastive method) and FOCAL~\cite{liu2023focal} (a modality-oriented contrastive method) as representative methods.

\textbf{Hybrid Paradigm.}
Hybrid paradigm combines multiple SSL paradigms to enhance representation learning. 
Existing methods primarily leverage reconstruction and contrastive objectives. %enabling the model to capture both local and global patterns of sensor data.
We select CrossHAR~\cite{hong2024crosshar}, a method that integrates masked reconstruction and augmentation-based contrastive learning specifically designed for generalizable HAR, and SimMTM~\cite{dong2023simmtm}, a manifold-based hybrid time-series method that demonstrates strong cross-domain generalization capability.
Detailed descriptions of the selected methods are provided in Appendix~\ref{app:method}. 

\subsection{Benchmark Setup}
\label{sec:setup}

\textbf{Task Setting.}
We consider eight target activity classes: \textit{sitting, standing, lying, upstairs, downstairs, walking, running}, and \textit{jumping}. 
These activities are selected for two reasons: 
(i) These activities are the most commonly shared activities across the 14 datasets (accounting for approximately two-thirds of the total samples), enabling comprehensive cross-dataset evaluation. 
(ii) These activities are representative of real-world HAR applications~\cite{xu2025experience, xu2025experience, straczkiewicz2021systematic, althoff2025countrywide}.
Unless otherwise specified, all experiments are conducted using samples from these eight activity classes.

\textbf{Split Setting.}
To conduct comprehensive cross-dataset evaluations, we adopt a dataset-level five-fold cross-validation protocol. 
Specifically, all datasets are partitioned into five disjoint groups. 
In each split, four groups are used as the \textit{source datasets} for training, while the remaining group serves as the \textit{target datasets} for evaluation. 
The target datasets are strictly held out and used for testing.
This setting allows models to learn motion patterns from multiple heterogeneous source datasets to improve generalization while evaluating performance on multiple unseen datasets with diverse data distributions.
Accuracy and F1-score are reported in our benchmark.

To systematically quantify the impact of domain shifts in sensor-based HAR, we conduct a series of experiments along different dimensions using the curated dataset. 
Specifically, we group the data according to three key factors:
(i) \textit{Cross-subject.} 
We perform subject-level five-fold cross-validation to evaluate the effect of subject-induced domain shifts on generalization.
(ii) \textit{Cross-location.} 
We divide the data into three body-location groups: trunk, upper limb, and lower limb, and conduct cross-validation across these groups to assess the impact of sensor placement variability.
(iii) \textit{Cross-device.} 
We partition the data into two groups according to device type: custom-grade devices (daily devices, such as smartphones and smartwatches) and research-grade devices (dedicated IMU data acquisition devices), to evaluate the effect of device-related domain shifts.
Detailed training and split configurations are provided in Appendix~\ref{app:training} and~\ref{app:split}.

\textbf{Architecture Setting.}
In our benchmark, SSL methods are first applied to train a sensor \textit{encoder} using unlabeled data with the goal of learning generalizable representations. 
A \textit{classifier} is then trained in a supervised manner on labeled data to map these representations to activity classes, while the encoder is frozen.
To ensure a fair and comprehensive evaluation, we adopt unified and diverse model backbones.
For the sensor encoder, we benchmark three widely used backbone architectures: Transformer, ResNet, and Convolutional Neural Network (CNN). 
For the classifier, we benchmark four backbone architectures: CNN, Multi-Layer Perceptron (MLP), Gated Recurrent Unit (GRU) and Transformer.
All of these backbones have been widely applied in sensor-based HAR studies~\cite{hong2024crosshar, liu2023focal, zhang2025mopformer, xu2025relcon}.
Detailed backbone architectures are provided in Appendix~\ref{app:architectures}.

\textbf{Modality Setting.}
We evaluate two common modality settings: accelerometer-only data~\cite{xu2025relcon, logacjov2024self, haresamudram2022assessing} and combined accelerometer and gyroscope data~\cite{hong2024crosshar, liu2023focal, zhang2025mopformer} to assess the impact of different modality combinations on generalizable sensor-based HAR.

%% file: 4-experiments.tex
\section{Results}
\label{sec:results}

\subsection{Benchmark Results}
\label{sec:model}

In this section, we summarize the main benchmark results, including SSL paradigms, methods, and model architectures. 
Additional results can be found in Appendix~\ref{app:model}.

\input{tables/overall}

\textbf{Overall Performance.}
We report the F1-scores of HAR performance in Table~\ref{tab:overall} with the mean and standard deviation obtained from dataset-level five-fold cross-validation. 
The accuracy results are provided in Appendix Table~\ref{tab:overall_acc}.
Overall, existing SSL methods remain insufficient under the cross-dataset generalization setting, with the best performance across all 192 results reaching 61.31\%.
Moreover, SSL methods specifically designed for HAR do not demonstrate clear advantages over general-purpose time-series SSL methods. 
These findings highlight the need for more advanced methods specifically designed to address generalization challenges in sensor-based HAR.
Across the four SSL paradigms, the hybrid paradigm achieves the best overall performance (median F1: 54.99\%), highlighting the effectiveness of combining reconstruction and contrastive SSL objectives for learning generalizable sensor representations.
Reconstruction and recognition paradigms also achieve comparable performance (median F1: 54.54\% and 53.34\%, respectively), while contrastive methods perform relatively worse (median F1: 49.19\%).
For modality combinations, models trained with accelerometer data outperform those using both accelerometer and gyroscope data, possibly because the selected activity classes rely primarily on accelerometer information~\cite{yamane2025impact}.

\textbf{SSL Methods.}
The contrastive method FOCAL achieves the best overall performance (average F1: 56.98\%), while the other contrastive methods, TS2Vec and TS-TCC, achieve the worst results among all eight methods (average F1: 48.37\% and 34.78\%, respectively), highlighting the importance of advanced contrastive learning design for generalizable HAR.
The hybrid method SimMTM also achieves competitive performance (average F1: 55.52\%) with FOCAL.
In contrast, the hybrid method CrossHAR and the reconstruction methods CRT and LIMU-BERT show comparable results (average F1: 53.40\%, 53.97\%, and 52.33\%, respectively).
Among reconstruction methods, CRT consistently outperforms LIMU-BERT, likely due to its frequency-domain modeling strategy that improves generalization in HAR, consistent with previous studies~\cite{napoli2024benchmark}.

\textbf{Encoder Architectures.}
For different encoder architectures, we observe that the CNN architecture achieves the best overall performance, with an average F1 of 53.01\%. 
This finding is consistent with previous results in both in-dataset~\cite{chan2024capture, qian2022makes} and small-scale cross-dataset~\cite{napoli2024benchmark} sensor-based HAR studies, further confirming the effectiveness of CNN in learning robust and generalizable representations from sensor data.
In addition, the ResNet encoder achieves comparable performance (average F1: 52.88\%).
The Transformer encoder performs notably worse (average F1: 47.08\%), likely due to the data scale is insufficient to fully exploit its representation learning capability.

\textbf{Classifier Architectures.}
Across different classifier architectures, we observe that CNN, GRU, and Transformer classifiers achieve comparable performance, with average F1 of 53.74\%, 51.97\%, and 51.94\%, respectively. 
In contrast, MLP performs significantly worse, reaching only 46.32\%.
This highlights that classifiers with more complex architectures are more effective than simpler architectures at leveraging SSL-learned representations for cross-dataset generalization. 
Meanwhile, among complex classifiers, the choice of specific architecture is not a primary factor that limits the generalization capability of sensor-based HAR models.

\subsection{Results Analysis}
\label{sec:analysis}

In this section, we investigate the impact of data scale and domain shift factors on the generalization performance of sensor-based HAR models.
Detailed evaluation configurations are provided in Appendix~\ref{app:additional}.
We address the following research questions (RQs).

\begin{figure}[t]
	\centering
	\includegraphics[width=5.2in]{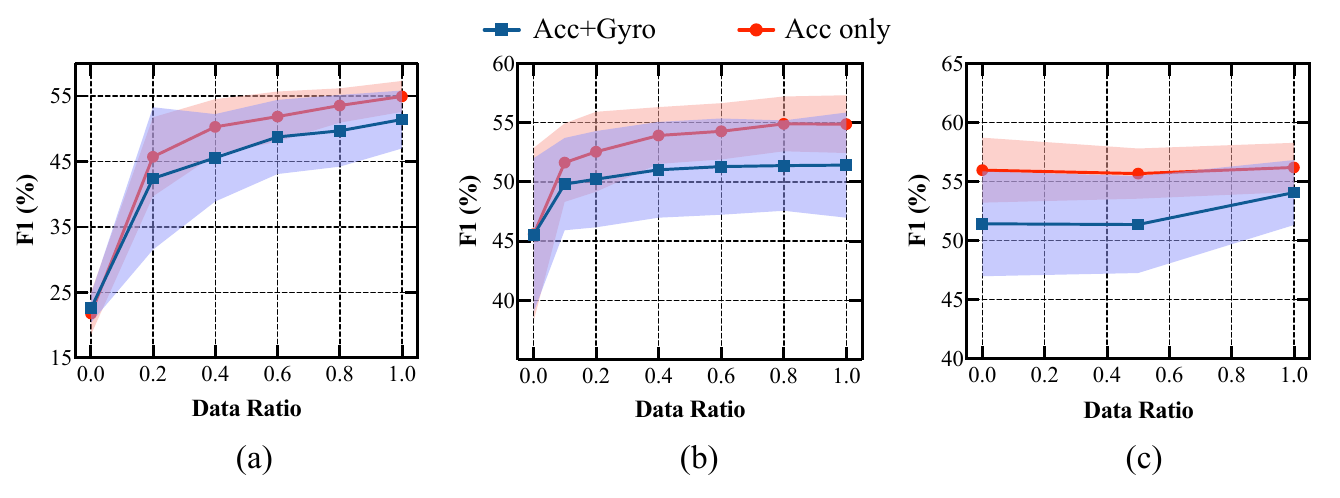}
	\caption{Impact of training data scale on generalization performance.
    Mean ± standard deviation of F1-score (\%) from four representative methods are reported.
    (a) Impact of encoder pretraining data size.
    (b) Impact of classifier training data size.
    (c) Impact of irrelevant activity data size.} 
	\label{fig:data}
\end{figure}

\textbf{RQ1: How does the unlabeled data scale affect sensor-based HAR generalization?}
 
To explore the impact of unlabeled data, we keep the classifier training data fixed while progressively scaling the amount of unlabeled pretraining data. 
As shown in Figure~\ref{fig:data} (a), increasing the amount of unlabeled pretraining data consistently improves the generalization capability of HAR models. 
\textit{This finding highlights the importance of integrating multiple datasets and collecting large-scale unlabeled sensor data to enhance the generalization capability of SSL HAR models.}

\textbf{RQ2: How does the labeled data scale affect sensor-based HAR generalization?}

To explore the impact of labeled data, we keep the pretraining data fixed while progressively increasing the amount of labeled data for classifier training. 
As shown in Figure~\ref{fig:data} (b), increasing the proportion of labeled data from 1\% to 10\% leads to substantial performance gains. 
However, further increasing the labeled data size does not yield additional improvements. 
\textit{This suggests that the generalization capability of SSL models is NOT primarily constrained by the availability of labeled data.}

\textbf{RQ3: Will pretraining with irrelevant activities benefit HAR generalization?}

It is important to investigate whether incorporating irrelevant sensor data for pretraining can improve generalization in HAR. 
To this end, we keep both the existing unlabeled and labeled data fixed, and progressively incorporate samples from activity classes outside the eight target classes into the unlabeled pretraining data.
As shown in Figure~\ref{fig:data} (c), we find that incorporating irrelevant activity samples does not lead to generalization performance improvements. 
\textit{
This suggests that expanding pretraining data size by increasing activity diversity without alignment to the target activity classes does not necessarily benefit generalization capability, highlighting the importance of task-relevant activity classes in pretraining.}
Detailed accuracy and F1-score results of the data scale for each method are provided in Appendix~\ref{app:data1}.

\textbf{RQ4: How do diverse domain shift factors affect sensor-based HAR generalization?}

\input{tables/cross}

We evaluate the impact of three major domain shift factors in sensor-based HAR: \textit{cross-subject}, \textit{cross-location}, and \textit{cross-device}. 
The results are reported in Table~\ref{tab:cross}.
For \textit{cross-subject} evaluation, we observe notably better performance compared to the overall results in Table~\ref{tab:overall}.
\textit{
This highlights the effectiveness of subject-level access to target dataset distributions in mitigating domain shift, while also suggesting that the widely adopted subject-level split evaluation may overestimate model robustness and fail to faithfully reflect real-world deployment scenarios.}
For \textit{cross-device} evaluation, we find that models trained on consumer-grade devices generalize better to research-grade devices. 
For \textit{cross-location} evaluation, we observe that data collected from the limb generalizes more effectively to trunk locations than vice versa.
\textit{These findings highlight the importance of data collected from consumer-grade devices and limb locations for developing generalizable sensor-based HAR systems.}
Detailed accuracy and F1-score results for each method are provided in Appendix~\ref{app:cross}.

%% file: tables/overall.tex
\begin{table}[t]
\centering
\footnotesize
\caption{Overall comparison results. 
Mean ± standard deviation of F1-score (\%) from dataset-level five-fold cross-validation are reported.
\textbf{Bold} indicates the best result within each row, and \underline{underline} indicates the best result within each column.}
\scalebox{0.68}{
\begin{tabular}{@{}c|c|c|c|cc|ccc|cc@{}}
\toprule
\multirow{2}{*}{Encoder} 
& \multirow{2}{*}{Classifier} 
& \multirow{2}{*}{Modality} 
& \multicolumn{1}{c|}{Recognition} 
& \multicolumn{2}{c|}{Reconstruction} 
& \multicolumn{3}{c|}{Contrastive} 
& \multicolumn{2}{c}{Hybrid} \\
\cmidrule(l){4-11}
& & 
& BioBankSSL 
& LIMU-BERT & CRT 
& TS-TCC & TS2Vec & FOCAL 
& SimMTM & CrossHAR \\ 
\midrule \midrule

\multirow{8}{*}{Transformer} 
& \multirow{2}{*}{CNN} & Acc & 53.26 ± 5.50 & 57.16 ± 3.26 & 57.07 ± 2.84 & 28.47 ± 9.87 & 53.75 ± 4.92 & 55.84 ± 5.12 & \textbf{58.87 ± 4.76} & 57.34 ± 2.22 \\
& & Acc+Gyro & 48.06 ± 9.27 & 54.98 ± 5.27 & \textbf{55.72 ± 4.98} & 21.36 ± 5.48 & 47.84 ± 6.62 & 49.59 ± 5.67 & 54.43 ± 4.05 & 54.21 ± 6.25 \\ 
\cmidrule(l){2-11}

& \multirow{2}{*}{MLP} & Acc & 53.69 ± 5.46 & 45.03 ± 5.84 & 50.67 ± 4.21 & 18.22 ± 2.16 & 22.43 ± 4.36 & 53.08 ± 4.22 & \textbf{53.73 ± 4.24} & 44.80 ± 2.67 \\
& & Acc+Gyro & 49.82 ± 9.46 & 39.85 ± 12.35 & 35.15 ± 19.16 & 19.25 ± 4.41 & 22.03 ± 6.99 & 50.17 ± 9.16 & \textbf{51.38 ± 4.62} & 40.31 ± 6.01 \\
\cmidrule(l){2-11}

& \multirow{2}{*}{GRU} & Acc & 54.06 ± 5.40 & 56.17 ± 3.78 & \textbf{57.26 ± 2.99} & 20.80 ± 2.03 & 50.88 ± 3.09 & 53.27 ± 4.00 & 54.62 ± 5.83 & 56.79 ± 5.03 \\
& & Acc+Gyro & 48.87 ± 9.76 & 52.71 ± 7.30 & \textbf{56.41 ± 6.10} & 20.15 ± 4.87 & 46.23 ± 7.57 & 48.30 ± 6.42 & 50.63 ± 6.26 & 52.82 ± 5.67 \\
\cmidrule(l){2-11}

& \multirow{2}{*}{Transformer} & Acc & 54.14 ± 5.12 & 54.65 ± 3.96 & \textbf{58.22 ± 2.77} & 21.49 ± 2.02 & 47.86 ± 3.03 & 52.62 ± 4.68 & 54.83 ± 3.85 & 56.58 ± 3.07 \\
& & Acc+Gyro & 49.01 ± 8.64 & 51.54 ± 6.89 & \textbf{57.83 ± 4.71} & 22.15 ± 5.35 & 43.14 ± 9.61 & 47.97 ± 7.13 & 50.89 ± 5.38 & 52.92 ± 5.18 \\

\midrule

\multirow{8}{*}{ResNet} 
& \multirow{2}{*}{CNN} & Acc & \textbf{\underline{61.31 ± 5.58}} & \underline{59.28 ± 4.05} & 53.83 ± 3.76 & 44.86 ± 2.91 & \underline{58.59 ± 4.40} & \underline{61.16 ± 3.74} & 58.76 ± 4.88 & \underline{58.54 ± 4.69} \\
& & Acc+Gyro & 54.09 ± 6.73 & 55.21 ± 7.05 & 58.10 ± 5.11 & 43.50 ± 8.12 & 54.12 ± 5.61 & \textbf{60.17 ± 4.48} & 57.63 ± 6.04 & 54.38 ± 5.97 \\
\cmidrule(l){2-11}

& \multirow{2}{*}{MLP} & Acc & 56.79 ± 4.88 & 44.17 ± 1.88 & 35.27 ± 9.71 & 40.79 ± 1.39 & 48.69 ± 2.52 & \textbf{59.72 ± 3.49} & 56.24 ± 5.14 & 51.04 ± 4.11 \\
& & Acc+Gyro & 49.69 ± 9.85 & 44.42 ± 6.18 & 51.18 ± 9.23 & 32.26 ± 10.16 & 49.24 ± 8.41 & \textbf{61.01 ± 4.57} & 55.99 ± 3.99 & 49.67 ± 6.66 \\
\cmidrule(l){2-11}

& \multirow{2}{*}{GRU} & Acc & 57.54 ± 5.24 & 56.39 ± 3.90 & 52.76 ± 4.89 & 42.48 ± 1.91 & 56.08 ± 4.33 & 58.66 ± 3.13 & \textbf{\underline{58.95 ± 4.78}} & 57.21 ± 6.02 \\
& & Acc+Gyro & 52.93 ± 5.54 & 52.38 ± 5.54 & 58.50 ± 5.29 & 33.04 ± 10.53 & 50.22 ± 6.37 & \textbf{59.69 ± 3.83} & 54.33 ± 5.09 & 52.55 ± 8.93 \\
\cmidrule(l){2-11}

& \multirow{2}{*}{Transformer} & Acc & \textbf{59.96 ± 6.39} & 57.05 ± 3.36 & 55.34 ± 3.82 & 41.01 ± 1.69 & 56.11 ± 4.25 & 59.76 ± 3.02 & 58.51 ± 4.71 & 55.55 ± 3.35 \\
& & Acc+Gyro & 51.56 ± 6.60 & 51.63 ± 5.11 & \textbf{\underline{59.91 ± 4.77}} & 31.74 ± 9.01 & 50.53 ± 6.10 & 58.96 ± 3.66 & 51.72 ± 5.12 & 51.35 ± 7.08 \\

\midrule

\multirow{8}{*}{CNN} 
& \multirow{2}{*}{CNN} & Acc & 55.35 ± 5.45 & 58.57 ± 3.79 & 55.89 ± 2.43 & \underline{47.82 ± 4.75} & 56.62 ± 3.08 & \textbf{58.78 ± 1.50} & 58.68 ± 5.01 & 57.66 ± 5.05 \\
& & Acc+Gyro & 49.78 ± 6.42 & 55.03 ± 5.19 & 57.26 ± 6.04 & 43.57 ± 4.94 & 53.07 ± 3.77 & \textbf{59.48 ± 4.15} & 55.46 ± 6.33 & 54.90 ± 4.37 \\
\cmidrule(l){2-11}

& \multirow{2}{*}{MLP} & Acc & 49.74 ± 5.03 & 44.55 ± 1.15 & 48.17 ± 5.97 & 43.55 ± 3.01 & 44.18 ± 4.06 & \textbf{58.14 ± 2.16} & 57.22 ± 4.28 & 50.99 ± 3.80 \\
& & Acc+Gyro & 43.73 ± 6.62 & 46.05 ± 5.51 & 50.60 ± 5.86 & 42.42 ± 5.13 & 41.80 ± 6.02 & \textbf{60.88 ± 4.04} & 55.50 ± 4.99 & 50.09 ± 8.58 \\
\cmidrule(l){2-11}

& \multirow{2}{*}{GRU} & Acc & 54.61 ± 4.67 & 57.15 ± 4.99 & 56.08 ± 2.92 & 46.55 ± 5.02 & 53.66 ± 3.92 & \textbf{59.27 ± 2.88} & 57.65 ± 4.17 & 57.49 ± 5.19 \\
& & Acc+Gyro & 48.56 ± 7.57 & 51.74 ± 4.63 & 57.41 ± 6.61 & 41.69 ± 7.08 & 50.45 ± 4.23 & \textbf{61.04 ± 5.47} & 53.85 ± 6.20 & 53.44 ± 5.82 \\
\cmidrule(l){2-11}

& \multirow{2}{*}{Transformer} & Acc & 55.42 ± 4.37 & 56.93 ± 3.14 & 57.53 ± 1.93 & 46.30 ± 4.67 & 54.09 ± 3.14 & \textbf{59.71 ± 2.34} & 57.83 ± 5.31 & 57.27 ± 3.87 \\
& & Acc+Gyro & 49.75 ± 6.19 & 53.36 ± 4.23 & 59.09 ± 4.96 & 41.27 ± 5.16 & 49.18 ± 3.91 & \textbf{60.20 ± 4.70} & 54.86 ± 6.06 & 53.74 ± 5.25 \\

\midrule
\bottomrule
\end{tabular}
}
\label{tab:overall}
\end{table}

%% file: tables/cross.tex
\begin{table}[t]
\centering
\footnotesize
\caption{Cross-domain performance. Mean ± standard deviation of accuracy and F1-score (\%) from four representative methods are reported.}
\begin{tabular}{c|c|c|cc|cc}
\toprule
\multirow{2}{*}{Setting}      &\multirow{2}{*}{Source}&\multirow{2}{*}{Target}& \multicolumn{2}{c|}{Acc-only}  & \multicolumn{2}{c}{Acc-gyro}  \\ \cmidrule(l){4-7} 
                              &             &             & Acc           & F1            & Acc           & F1            \\ \midrule
Cross-user                    & -           & -           & 70.57 ± 3.79  & 70.27 ± 4.24  & 72.07 ± 1.13  & 73.12 ± 1.32 \\ \midrule 
\multirow{2}{*}{Cross-device} & Custom      & Research    & 55.71 ± 2.00  & 55.87 ± 2.62  & 57.02 ± 3.60  & 56.29 ± 4.62  \\
                              & Research    & Custom      & 52.74 ± 2.16  & 50.76 ± 1.79  & 50.38 ± 4.69  & 48.32 ± 5.05  \\ \midrule
\multirow{3}{*}{Cross-location}& Trunk      & Upper limb  & 48.93 ± 2.87  & 49.47 ± 3.79  & 51.96 ± 2.05  & 52.48 ± 1.97 \\
                              & Trunk       & Lower limb  & 40.05 ± 3.56  & 38.30 ± 3.87  & 43.05 ± 6.67  & 42.96 ± 6.58 \\
                              & Upper limb  & Trunk       & 49.61 ± 1.45  & 49.57 ± 1.79  & 48.25 ± 3.88  & 47.73 ± 4.68 \\
                              & Upper limb  & Lower limb  & 43.56 ± 2.95  & 42.59 ± 3.85  & 43.76 ± 6.71  & 42.52 ± 6.89  \\ 
                              & Lower limb  & Trunk       & 46.10 ± 2.56  & 44.35 ± 3.35  & 48.52 ± 2.99  & 47.81 ± 4.46  \\
                              & Lower limb  & Upper limb  & 46.53 ± 1.74  & 43.95 ± 3.24  & 50.28 ± 2.83  & 47.70 ± 3.73  \\\bottomrule 
\end{tabular}
\label{tab:cross}
\end{table}

%% file: 5-discussion.tex
\section{Limitations and Future Works}
\label{sec:limitations}

Several limitations of this work remain and warrant further investigation.
First, the downstream HAR task in $\mathsf{BenchHAR}$ focuses on eight activity classes from 14 datasets, primarily due to discrepancies in activity classes across datasets. 
Expanding data scale and activity class coverage for cross-dataset HAR remains an important direction for future work.
Second, this benchmark focuses on evaluating SSL methods for HAR based on IMU sensors, serving as a starting point for multimodal sensor foundation models. 
Future work can extend to multimodal wearable data, as well as explore scaling laws to further improve generalization capability~\cite{narayanswamy2025scaling}.

%% file: 6-conclusion.tex
\section{Conclusion}
\label{sec:conclusion}

In this paper, we present $\mathsf{BenchHAR}$, a comprehensive benchmark for evaluating the cross-dataset generalization capability of SSL methods in sensor-based HAR. 
By establishing this benchmark, we aim to facilitate research on generalizable HAR and provide valuable resources and useful insights for developing more robust and effective learning approaches toward real-world deployment.

%% file: 7-appendix.tex
\newpage
\appendix

% \section*{Appendix}

\section{Benchmark Dataset}
\label{app:dataset}

\subsection{Datasets Overview}
\label{app:overview}

We detail the information for the 14 datasets employed in our benchmark.
As we focus on both accelerometer and gyroscope data, we exclude datasets that only provide accelerometer data, such as Capture-24~\cite{chan2024capture} and HAR70+~\cite{ustad2023validation}.
The dataset statistics are summarized in Table~\ref{tab:datasets}.

\textbf{UCI}~\cite{reyes2016transition} includes 30 volunteers aged 19 to 48. 
Data are collected using a smartphone worn at the waist. 
The dataset covers six activities of daily living, including standing, sitting, lying, walking, walking downstairs, and walking upstairs, together with postural transitions. 
Each subject performs two trials: in the first trial, the smartphone was fixed on the left side of the waist belt. 
In the second trial, the phone position is adjusted freely by the subject.

\textbf{HHAR}~\cite{stisen2015smart} consists of data collected from 9 users using a combination of 4 smartwatches and 8 smartphones in unconstrained, real-world settings. 
Participants perform a set of predefined activities in random order, including biking, sitting, standing, walking, walking upstairs, and walking downstairs. 

\textbf{Shoaib}~\cite{shoaib2014fusion} consists of data collected from 10 participants aged 25 to 30 years. 
Each participant is equipped with five smartphones on five different body positions: right jean’s and left jean’s pockets, belt, right upper arm and right wrist.
The dataset covers seven physical activities: walking, jogging, sitting, standing, biking, walking upstairs, and walking downstairs. 

\textbf{Motion}~\cite{malekzadeh2019mobile} consists of data collected from 24 participants using a smartphone placed in the front pocket. 
The dataset includes six activities: walking, jogging, sitting, standing, walking upstairs, and walking downstairs. 
The participants range in age from 18 to 46 years (mean = 29), with diverse physical characteristics (height: 161-190 cm, weight: 48-102 kg) and a balanced gender distribution (14 male, 10 female).

\textbf{DSADS}~\cite{barshan2014recognizing} consists of 18 daily and sports activities collected from eight subjects (4 female and 4 male), aged between 20 and 30.
Each subject performs activities while wearing multiple sensing devices placed on multiple body locations: torso, right arm, left arm, right leg, and left leg.

\textbf{USC-HAD}~\cite{zhang2012usc} contains 12 daily activities collected from 14 subjects (7 male and 7 female), aged between 21 and 49.
Each subject wears a single sensing device placed at the front-right hip.

\textbf{KU-HAR}~\cite{sikder2021ku} contains 18 daily and sports activities collected from 90 participants (75 male and 15 female).
Data are recorded using smartphone sensors worn at the waist.

\textbf{PAMAP2}~\cite{reiss2012introducing} contains 18 different physical activities collected from nine subjects, such as walking, cycling, and playing soccer.
Data are collected using IMU sensors positioned on the wrist, chest, and ankle, together with a heart rate monitor.
We only use the IMU data in our benchmark.

\textbf{TNDA-HAR}~\cite{liao2022deep} contains eight activities, including three static activities and 5 periodic activities (walking, running, cycling, and walking upstairs/downstairs), collected from 50 subjects.
The wearable sensors are positioned on the left ankle, left knee, back, right wrist, and right arm.

\textbf{Mhealth}~\cite{banos2014mhealthdroid} contains 12 activities, including daily and sports activities, such as standing, climbing, and knee bending.
Data are collected from 10 subjects.
The sensors are positioned on the chest, right wrist, and left ankle.

\textbf{WISDM}~\cite{wisdm} contains 18 activities (such as walking, clapping, and folding clothes) collected from 51 subjects.
Data are recorded using a smartphone in the pocket and a smartwatch on the hand.

\textbf{RealWorld}~\cite{sztyler2016body} contains 8 activities (walking, running, sitting, standing, lying, stairs up, stairs down, and jumping) collected from 15 subjects (8 male and 7 female).
Sensors are positioned on the chest, forearm, head, shin, thigh, upper arm, and waist, recording acceleration, gyroscope, GPS, light, magnetic field, and sound level data.

\textbf{HARSense}~\cite{harsense} contains 6 activities (walking, standing, upstairs, downstairs, running, and sitting) collected from 12 subjects aged above 23.
Sensors are placed at the waist and in the front pockets.

\textbf{UT-Complex}~\cite{shoaib2016complex} contains 13 activities, including daily activities such as walking, jogging, cycling, smoking, and talking, collected from 10 male subjects aged between 23 and 35.
Smartphones are placed in the pocket and on the right wrist.

\input{tables/datasets}

\subsection{Benchmark Statistics}
\label{app:statistics}

\begin{figure}[t]
	\centering
	\includegraphics[width=3.8in]{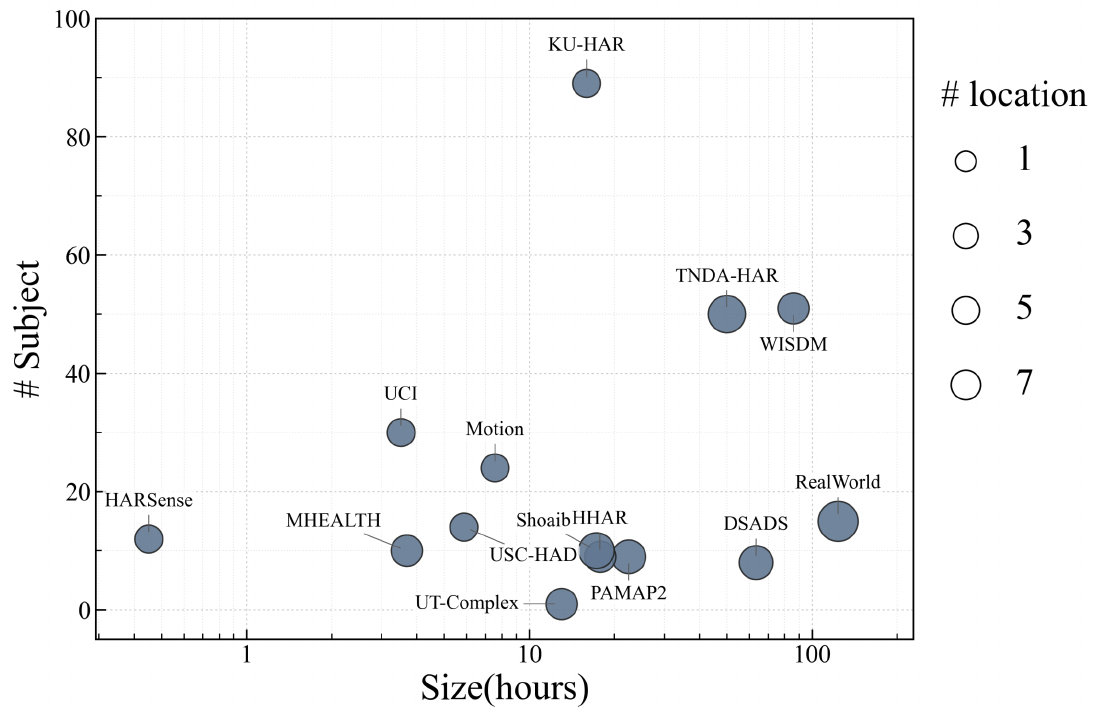}
	\caption{Distribution of the 14 sensor-based HAR datasets.} 
	\label{fig:vis}
\end{figure}

We provide detailed statistics of the standardized large-scale dataset.
Table~\ref{tab:activities} presents the 62 standardized activity classes across the 14 datasets.
Among them, \textit{sitting, standing, lying, upstairs, downstairs, walking, running}, and \textit{jumping} are the most commonly shared activities across datasets, while the remaining activities exhibit very limited overlap.
This highlights the substantial heterogeneity in the activity composition among different HAR datasets.
Figure~\ref{fig:vis} shows the distribution of the 14 datasets after standardization.
KU-HAR contributes the largest number of subjects, while RealWorld provides the largest amount of data.

\input{tables/activities}

\section{Method Description}
\label{app:method}

\subsection{Recognition Paradigm}

\textbf{BioBankSSL}~\cite{yuan2024self} represents a recent method of the SSL recognition paradigm for sensor data. 
The method constructs pretext tasks by randomly applying one or multiple transformations to each sensor sample, and the pretraining objective is to recognize whether a given transformation has been applied~\cite{saeed2019multi}. 
A broad set of transformation strategies has been explored in HAR~\cite{khowaja2024refuseact, saeed2019multi}, we follow the original BioBankSSL implementation and adopt three representative transformations: \textit{arrow of time}, \textit{permutation}, and \textit{time warping}. 
BioBankSSL demonstrates strong performance under large-scale pretraining~\cite{yuan2024self}, and thus serves as a representative SOTA approach of recognition-based SSL paradigms for sensor-based HAR.

\subsection{Reconstruction Paradigm}

\textbf{LIMU-BERT}~\cite{xu2021limu} extends the BERT-style masked reconstruction paradigm~\cite{devlin2019bert} to IMU sensor data. 
Instead of masking individual time steps, the method applies a binary masking strategy over contiguous subsequences, enabling the model to capture temporal dependencies and contextual structures in sensor data. 
Following the original implementation, we apply a masking ratio of 15\% to each channel across all IMU samples. 
The pretraining objective is to reconstruct the masked portions of the sequence. 
LIMU-BERT is one of the most widely adopted approaches in sensor-based HAR and has been successfully deployed in large-scale real-world applications~\cite{xu2025experience}.

\textbf{CRT}~\cite{zhang2023self} employs a dropping-based reconstruction strategy. 
Specifically, random contiguous blocks of the input are randomly removed, making the reconstruction task increasingly challenging over time.
In addition, CRT jointly leverages both time-domain and frequency-domain representations, requiring the model to reconstruct the dropped patches across these complementary views. 
This design introduces an alternative paradigm to reconstruction SSL reconstruction by explicitly modeling missing information through stochastic drop, while incorporating frequency-domain cues that are beneficial for capturing motion patterns in sensor data~\cite{napoli2024benchmark}. 
CRT achieves strong generalization capability in sensor-based HAR.

\subsection{Contrastive Paradigm}

\textbf{TS-TCC}~\cite{ijcai2021-324} integrates the ideas of CPC~\cite{oord2018representation} and SimCLR~\cite{chen2020simple} for contrastive pretraining on time-series data.
The method generates two augmented views of each sample via strong and weak augmentations. 
It then performs 
(i) temporal contrasting, where the model predicts future representations of the weakly augmented sequence conditioned on past and present representations of the strongly augmented sequence, and (ii) contextual contrasting, which enforces consistency across representations from different temporal contexts. 
Following the original implementation, we adopt \textit{shuffling} and \textit{scaling} as augmentation strategies. 
TS-TCC has demonstrated strong performance on contrastive learning benchmarks for sensor-based HAR~\cite{qian2022makes}.

\textbf{TS2Vec}~\cite{yue2022ts2vec} is a widely adopted general time-series contrastive learning framework. 
The method first generates augmented views of the input sequence, and then performs: (i) temporal contrasting, where representations at the same timestamp across different views are treated as positive pairs while those from different timestamps serve as negative pairs, and (ii) hierarchical contrasting to capture multi-scale temporal dependencies and learn representations at different resolutions.
Following the original implementation, we adopt \textit{timestamp masking} and \textit{random cropping} as augmentation strategies. 
TS2Vec has demonstrated strong performance in sensor-based HAR task~\cite{lee2024soft}.

\textbf{FOCAL}~\cite{liu2023focal} is a multimodal contrastive learning framework for time-series sensor data. 
The method generates augmented views using a diverse set of time-domain and frequency-domain transformations. 
It then optimizes multiple objectives: 
(i) inter-modality contrastive learning to capture modality-shared information, 
(ii) intra-modality contrastive learning over augmented samples to learn modality-specific information, 
(iii) orthogonality constraints between shared and private features within the same modality, as well as across private features of different modalities, to enforce disentangled representations in a decomposed feature space, and 
(iv) information regularization via sequence ordering, which specifies coarse-grained relationships between intra-sequence and inter-sequence distances. 
FOCAL achieves strong performance across multiple time-series sensing tasks~\cite{liu2023focal}, including sensor-based HAR.

\subsection{Hybrid Paradigm}

\textbf{CrossHAR}~\cite{hong2024crosshar} is a recent method designed for cross-dataset generalization in sensor-based HAR.
The framework consists of two key components: 
(i) a physically informed sensor data augmentation strategy to expand training diversity, and 
(ii) a hierarchical pretraining scheme that jointly leverages masked reconstruction and contrastive learning. 
Specifically, the model is first updated using a masked reconstruction objective, and subsequently optimized with a combination of reconstruction and contrastive objectives to enhance representation robustness. 
Following the original implementation, we apply a masking ratio of 15\% over subsequences for each channel across all IMU samples, and adopt \textit{shuffling} and \textit{scaling} as augmentation strategies for contrastive learning. 
For fair comparisons, we remove the physically informed data augmentation component in our implementation.

\textbf{SimMTM}~\cite{dong2023simmtm} establishes a connection between masked reconstruction and manifold learning for time-series data. 
Instead of reconstructing masked points solely from local context, SimMTM recovers them by aggregating multiple off-manifold neighbors with adaptive weighting, effectively assembling corrupted yet complementary temporal variations from different masked sequences to simplify the reconstruction task. 
To avoid trivial aggregation, SimMTM further incorporates a contrastive learning-based neighborhood assumption on the time-series manifold, encouraging the model to preserve meaningful local structure during reconstruction.
% SimMTM achieves strong cross-domain classification performance in sensor-based HAR task~\cite{liu2025robusthar}.
SimMTM achieves strong cross-domain classification performance in various general time-series tasks.

\section{Implementation Details}
\label{app:implementation}

\subsection{Training Details}
\label{app:training}

For all model configurations, we pretraining sensor encoders for 200 epochs, with a learning rate of 0.001 and a batch size of 512. 
For classifiers, training epochs are set to 100, with a learning rate of 0.001 and a batch size of 256. 
For all experiments, the training splits are divided into training and validation sets with a ratio of 8:2, while all samples in the test splits are used exclusively for evaluation.
All models are trained with the Adam optimizer on NVIDIA GeForce RTX 5090 GPUs.

\subsection{Data Split}
\label{app:split}

\input{tables/split}

The details of the dataset-level five-fold cross-validation are shown in Table~\ref{tab:split}. 
Each split contains 2--3 datasets, and we carefully partition the datasets to ensure that the amount of activity data is balanced across splits as much as possible.

\subsection{Model Architectures}
\label{app:architectures}

In this section, we describe the architectures of the encoders and classifiers in detail.
For other components in the selected methods, such as the decoders in reconstruction paradigm and the projectors in contrastive paradigm, we strictly follow the original implementations.

\subsubsection{Encoder}

\textbf{Convolutional Neural Network (CNN).}
We implement a temporal convolutional encoder with progressively increasing dilation factors to capture multi-scale temporal dependencies.
The encoder consists of 4 convolutional blocks with dilation factors $1$, $2$, $4$, and $8$.
Each block applies a 1D convolution with kernel size 5, followed by Group Normalization, GELU activation, and dropout with probability $p=0.1$:
$
\text{Conv1D} \rightarrow \text{GroupNorm} \rightarrow \text{GELU} \rightarrow \text{Dropout}.
$
The hidden dimension is set to $H=72$, and intermediate blocks expand the channel dimension by a factor of 2 before projecting back to the hidden space.
Specifically, the channel dimensions are
$
6 \rightarrow 144 \rightarrow 144 \rightarrow 144 \rightarrow 72.
$
A final $1\times1$ convolution is applied to project features into the fixed hidden dimension, followed by Group Normalization.
The overall architecture is:
$
[\text{ Input} \rightarrow \text{Transpose} \rightarrow
\text{ConvBlock}_{d=1}(6,144,k=5) \rightarrow
\text{ConvBlock}_{d=2}(144,144,k=5) \rightarrow
\text{ConvBlock}_{d=4}(144,144,k=5) \rightarrow
\text{ConvBlock}_{d=8}(144,72,k=5) \rightarrow
\text{1}\times\text{1 Conv}(72,72) \rightarrow \text{GroupNorm} \rightarrow
\text{Transpose} \rightarrow \text{Output }].
$

\textbf{ResNet.}
We adopt a ResNet-based encoder composed of multiple sequence-length-preserving residual blocks.
A stem layer first projects the input channels into a hidden space of dimension $H=72$ using a 1D convolution with kernel size 5, followed by GroupNorm and GELU activation.
The encoder then applies 4 residual blocks with dilation rates $1$, $2$, $4$, and $8$ to enlarge the temporal receptive field.
Each residual block contains two dilated 1D convolutional layers with kernel size 5, GroupNorm, GELU activation, and dropout with probability $p=0.1$, together with a skip connection.
Specifically, each block is defined as
$
\text{Conv1D} \rightarrow \text{GroupNorm} \rightarrow \text{GELU} \rightarrow \text{Dropout} \rightarrow
\text{Conv1D} \rightarrow \text{GroupNorm} \rightarrow \text{Dropout} \rightarrow
\text{Residual Add} \rightarrow \text{GELU}.
$
A final GroupNorm layer is applied before projecting the representation back to the sequence format.
The overall architecture is:
$
[\text{ Input} \rightarrow \text{Transpose} \rightarrow
\text{Conv1D}(\cdot,72,k=5,p=2) \rightarrow \text{GroupNorm} \rightarrow \text{GELU} \rightarrow
\text{ResBlock}_{d=1} \rightarrow \text{ResBlock}_{d=2} \rightarrow
\text{ResBlock}_{d=4} \rightarrow \text{ResBlock}_{d=8} \rightarrow
\text{GroupNorm} \rightarrow \text{Transpose} \rightarrow
\text{Output }].
$

\textbf{Transformer.}
We adopt a Transformer-based encoder for time-series modeling.
Each input token is first projected from the raw feature space to a hidden space of dimension $H=72$ through a linear layer, and then combined with learnable positional embeddings.
The encoder applies $L_t = 2$ stacked self-attention layers with 4 attention heads.
Each layer uses a position-wise feed-forward network with hidden dimension 144, together with residual connections and layer normalization.
Following the implementation, the same self-attention and feed-forward modules are shared across the stacked layers.
The overall architecture is:
$
[\text{ Input} \rightarrow
\text{Linear Projection}(72) \rightarrow
\text{Learnable Positional Embedding} \rightarrow
\text{Transformer Layer}(H=72,\text{heads}=4,\text{FFN}=144) \times 2 \rightarrow
\text{Output }].
$

\subsubsection{Classifier}

\textbf{Gated Recurrent Unit (GRU).}
For GRU-based classifier, the backbone consists of $L_r = 2$ stacked GRU blocks.
The first GRU block uses a hidden size of 20 with 2 recurrent layers, while the second GRU block uses a hidden size of 10 with 1 recurrent layer.
The hidden state at the last time step is used as the sequence representation, followed by dropout with probability $p=0.5$ and a fully connected layer for 3-class classification.
The overall architecture is:
$
[\text{ Input} \rightarrow \text{GRU}(\text{hidden}=20,\text{layers}=2)
\rightarrow \text{GRU}(\text{hidden}=10,\text{layers}=1)
\rightarrow \text{Last Time Step Selection}
\rightarrow \text{Dropout}(p=0.5)
\rightarrow \text{Linear}(10)
\rightarrow \text{Output }].
$

\textbf{Multi-Layer Perceptron (MLP).}
For MLP classifier, the input is first flattened into a single vector, followed by dropout with probability $p=0.5$.
The classifier contains $L_f = 2$ fully connected layers.
The first fully connected layer projects the flattened representation to a 64-dimensional hidden space, followed by a ReLU activation.
The second fully connected layer maps the hidden representation to the output space with dimension $C$, where $C$ denotes the number of target classes.
The overall architecture is:
$
[\text{ Input} \rightarrow
\text{Flatten} \rightarrow
\text{Dropout}(p=0.5) \rightarrow
\text{Linear}(64) \rightarrow
\text{ReLU} \rightarrow
\text{Dropout}(p=0.5) \rightarrow
\text{Linear}(64) \rightarrow
\text{Output }].
$

\textbf{Transformer.}
For Transformer-based classifier, the input sequence is first projected into a 64-dimensional latent embedding space and combined with learnable positional embeddings.
The encoder consists of $L_t = 1$ Transformer block with 4 attention heads and a feed-forward hidden dimension of 128.
The sequence representation is obtained by mean pooling over the temporal dimension, followed by dropout with probability $p=0.5$ and a final linear layer for classification.
The output dimension is $C$, where $C$ denotes the number of target classes.
The overall architecture is:
$
[\text{ Input} \rightarrow
\text{Linear Projection}(64) \rightarrow
\text{Positional Embedding} \rightarrow
\text{Transformer Block}(d=64,\text{heads}=4,\text{ff}=128) \times 1 \rightarrow
\text{Mean Pooling} \rightarrow
\text{Dropout}(p=0.5) \rightarrow
\text{Linear}(64) \rightarrow
\text{Output }].
$

\textbf{Convolutional Neural Network (CNN).}
For CNN-based classifier, the input is first transposed so that convolution is performed along the temporal axis.
The convolutional encoder consists of $L_c = 3$ convolutional blocks.
The three blocks use channel dimensions $6 \rightarrow 32$, $32 \rightarrow 64$, and $64 \rightarrow 64$, with kernel sizes $5$, $5$, and $3$, and paddings $2$, $2$, and $1$, respectively.
Each convolutional block is implemented as
$
\text{Conv1D} \rightarrow \text{ReLU} \rightarrow \text{GroupNorm} \rightarrow \text{MaxPool},
$
where the max-pooling layer uses kernel size 2, stride 2, and padding 0.
A global average pooling layer aggregates temporal information into a 64-dimensional representation.
This representation is then passed through dropout with probability $p=0.5$, followed by two fully connected layers.
The first fully connected layer maps the feature to a 64-dimensional hidden representation, followed by ReLU and dropout with probability $p=0.5$.
The second fully connected layer maps the hidden representation to the output space with dimension $C$, where $C$ denotes the number of target classes.
The overall architecture is:
$
[\text{ Input} \rightarrow
\text{Transpose} \rightarrow
\text{Conv1D}(6,32,k=5,p=2) \rightarrow \text{ReLU} \rightarrow \text{GroupNorm} \rightarrow \text{MaxPool}(2,2) \rightarrow
\text{Conv1D}(32,64,k=5,p=2) \rightarrow \text{ReLU} \rightarrow \text{GroupNorm} \rightarrow \text{MaxPool}(2,2) \rightarrow
\text{Conv1D}(64,64,k=3,p=1) \rightarrow \text{ReLU} \rightarrow \text{GroupNorm} \rightarrow \text{MaxPool}(2,2) \rightarrow
\text{Global AvgPool} \rightarrow
\text{Dropout}(p=0.5) \rightarrow
\text{Linear}(64,64) \rightarrow
\text{ReLU} \rightarrow
\text{Dropout}(p=0.5) \rightarrow
\text{Linear}(64,C) \rightarrow
\text{Output }].
$

\subsection{Implementation Details of Results Analysis}
\label{app:additional}

We select four SSL methods (BioBankSSL, CRT, FOCAL, and SimMTM) as they achieve the best performance within their respective paradigms in Section~\ref{sec:model}. 
For the data scale analysis (RQ1, RQ2 and RQ3), we employ Transformer as encoder and decoder architectures, as it is widely used in prior studies~\cite{narayanswamy2025scaling, ghorbani2022scaling, hoffmann2022an, shi2024scaling} to analyze the effect of data scale on model performance.
For the distribution shift factor analysis (RQ4), we adopt the encoder-classifier architecture combinations that achieve the best performance for the four methods in Section~\ref{sec:model}.

\newpage
\section{Additional Results}
\label{app:results}

In this section, we provide detailed evaluation results of Section~\ref{sec:results}.

\subsection{Benchmark Results}
\label{app:model}

Table~\ref{tab:overall_acc} reports the detailed accuracy results of HAR performance, including the mean and standard deviation obtained from dataset-level five-fold cross-validation. 
The corresponding F1-score results are provided in Table~\ref{tab:overall}. 
For accuracy, the best performance across all 192 results reaches 63.26\%.

\input{tables/acc}

\newpage
\subsection{Results Analysis}
\label{app:data}

\subsubsection{Impact of Data Scale}
\label{app:data1}

\begin{figure}[H]
	\centering
	\includegraphics[width=5.2in]{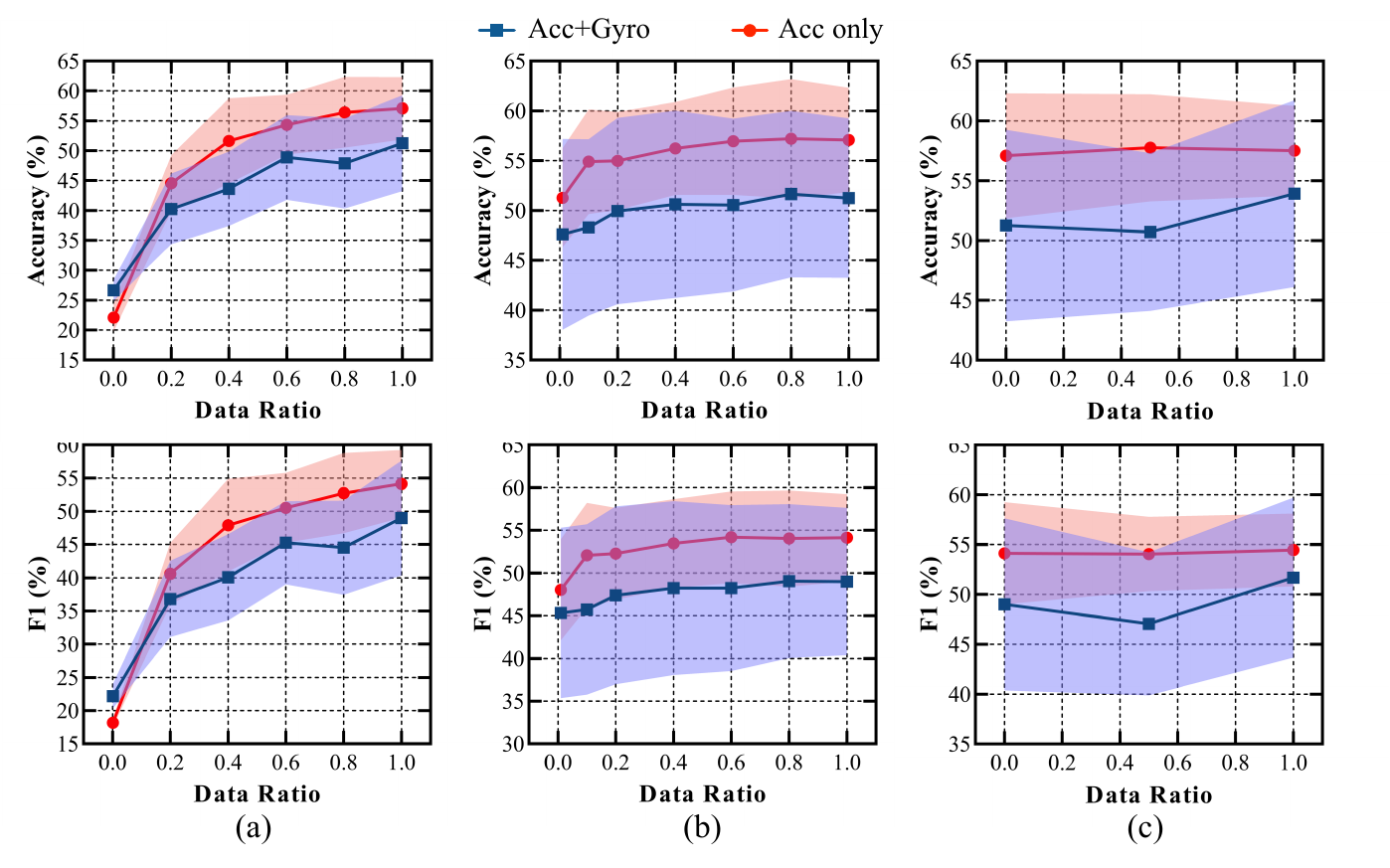}
	\caption{Impact of training data scale on the generalization performance of BioBankSSL.
    Mean ± standard deviation of accuracy and F1-score (\%) are reported.
    (a) Impact of encoder pretraining data size.
    (b) Impact of classifier training data size.
    (c) Impact of irrelevant activity data size.} 
	\label{fig:biobankssl}
\end{figure}

\begin{figure}[H]
	\centering
	\includegraphics[width=5.2in]{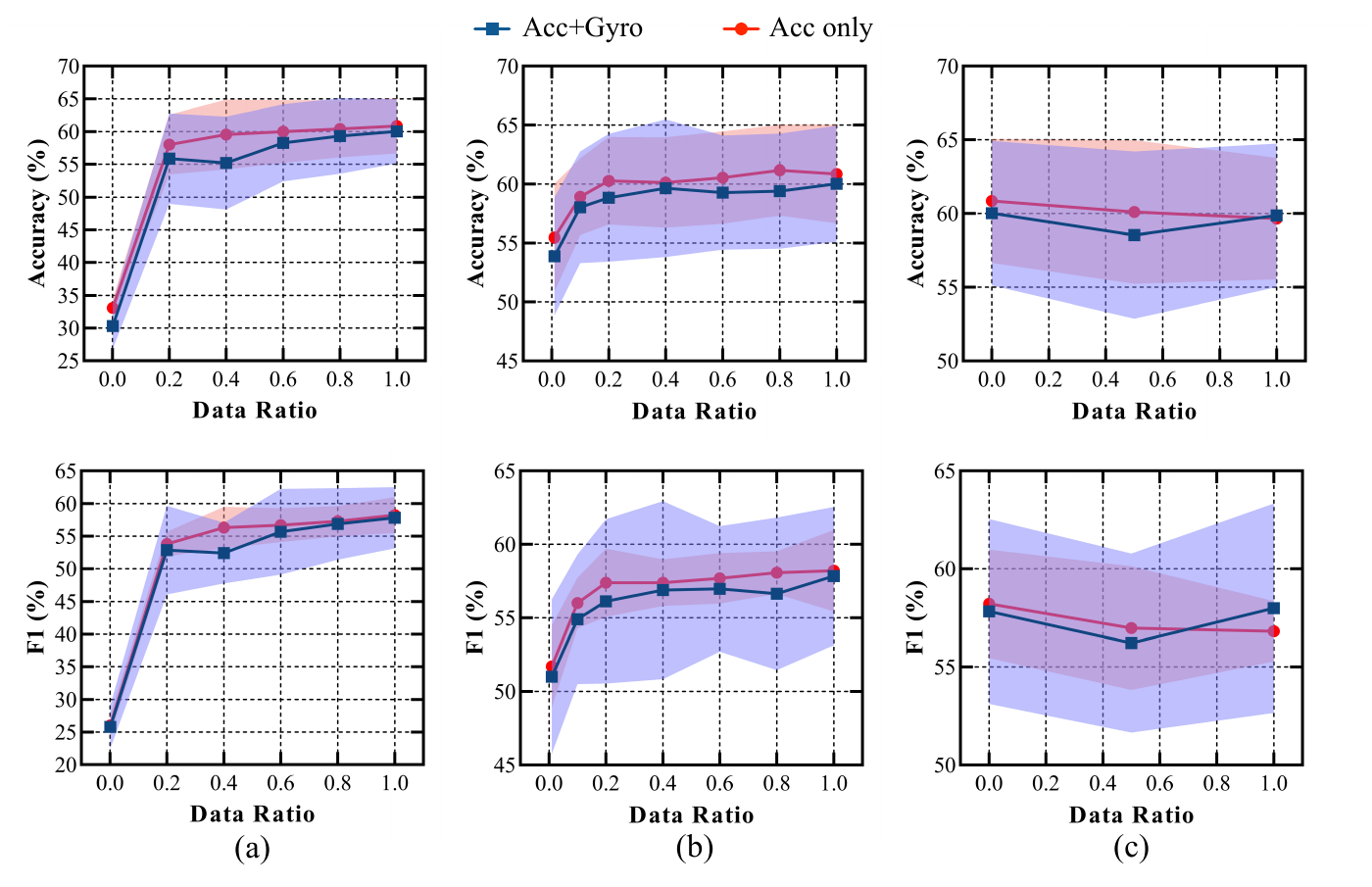}
	\caption{Impact of training data scale on the generalization performance of CRT.
    Mean ± standard deviation of accuracy and F1-score (\%) are reported.
    (a) Impact of encoder pretraining data size.
    (b) Impact of classifier training data size.
    (c) Impact of irrelevant activity data size.} 
	\label{fig:crt}
\end{figure}

\begin{figure}[H]
	\centering
	\includegraphics[width=5.2in]{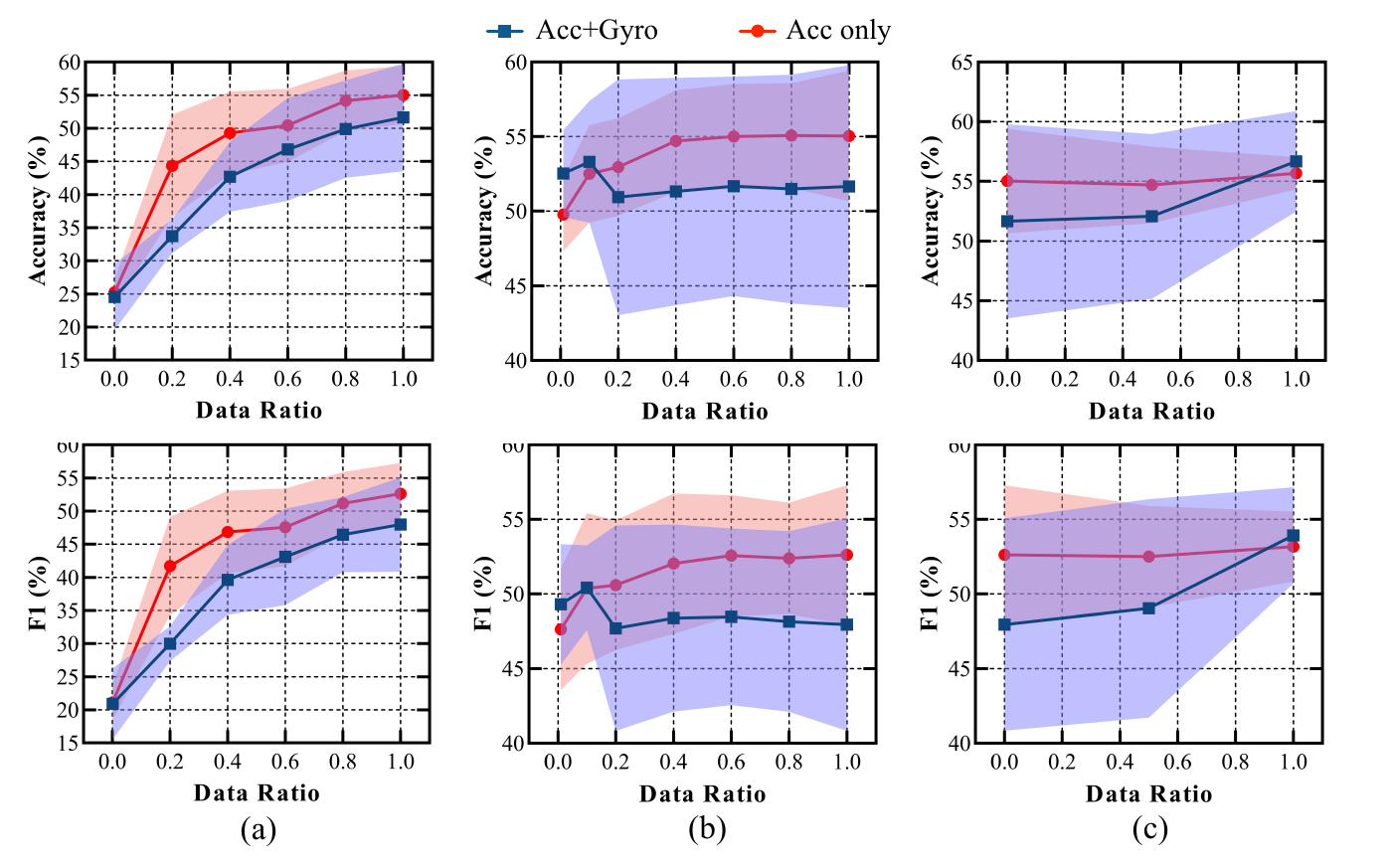}
	\caption{Impact of training data scale on the generalization performance of FOCAL.
    Mean ± standard deviation of accuracy and F1-score (\%) are reported.
    (a) Impact of encoder pretraining data size.
    (b) Impact of classifier training data size.
    (c) Impact of irrelevant activity data size.} 
	\label{fig:focal}
\end{figure}

\begin{figure}[H]
	\centering
	\includegraphics[width=5.2in]{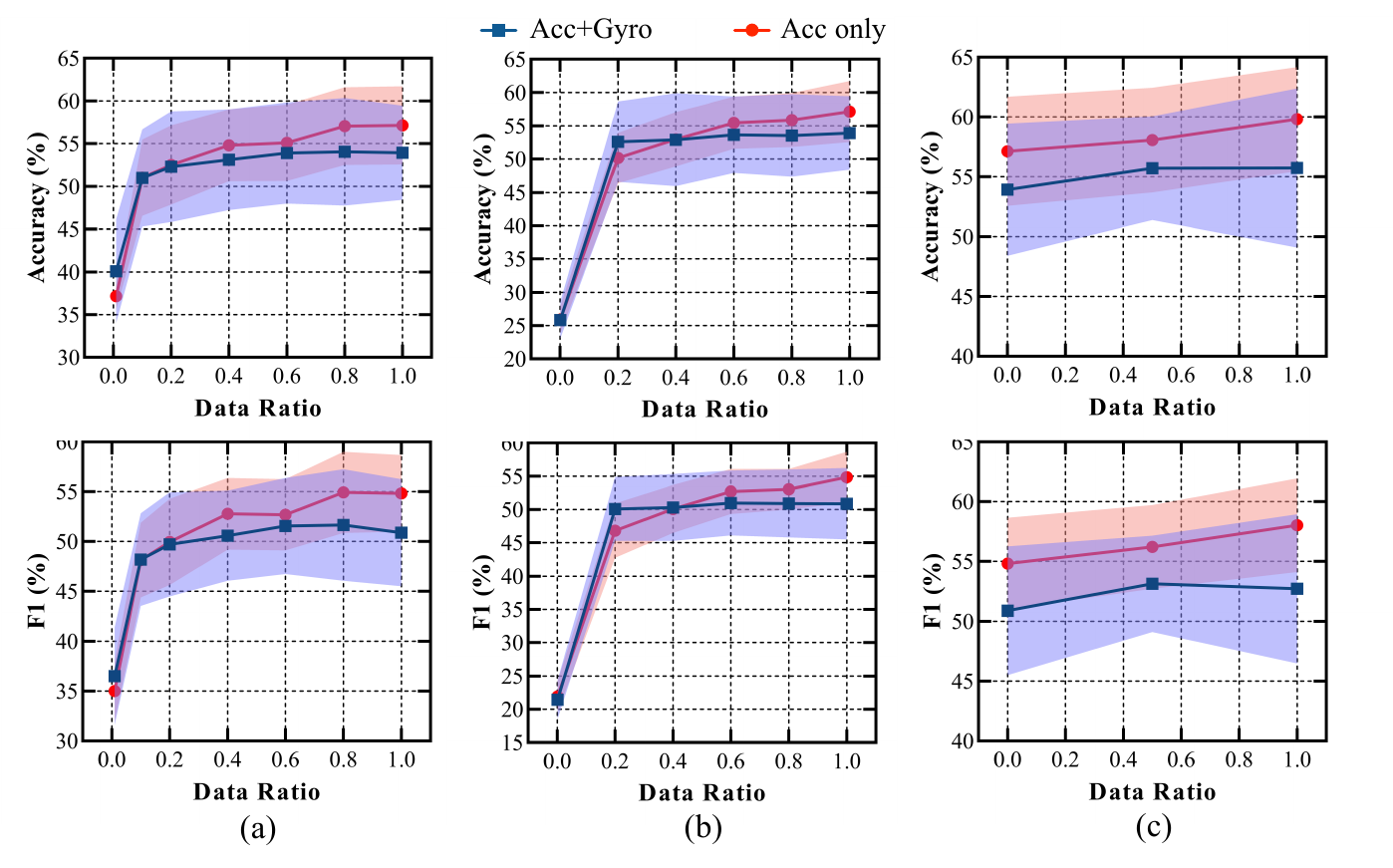}
	\caption{Impact of training data scale on the generalization performance of SimMTM.
    Mean ± standard deviation of accuracy and F1-score (\%) are reported.
    (a) Impact of encoder pretraining data size.
    (b) Impact of classifier training data size.
    (c) Impact of irrelevant activity data size.} 
	\label{fig:simmtm}
\end{figure}

\newpage
\subsubsection{Impact of Cross-domain Factors}
\label{app:cross}

We report detailed \textit{cross-subject}, \textit{cross-device} and \textit{cross-location} evaluation results in Tables~\ref{tab:subject}, \ref{tab:device}, and \ref{tab:location}, respectively. 
SimMTM achieves the best cross-subject performance, while CRT obtains the best performance in both the cross-device and cross-position settings.

\input{tables/cross-subject}

\input{tables/cross-device}

\input{tables/cross-location}

%% file: tables/datasets.tex
\definecolor{lightblue}{RGB}{220,235,250}

\begin{table}[h]
\centering
\footnotesize
\caption{Statistics of 14 datasets curated in our benchmark.}
\scalebox{0.97}{
\begin{tabular}{
>{\centering\arraybackslash}m{1.3cm}
>{\centering\arraybackslash}m{1.5cm}
>{\centering\arraybackslash}m{1.0cm}
>{\centering\arraybackslash}m{1.cm}
>{\centering\arraybackslash}m{1.1cm}
>{\centering\arraybackslash}m{1.cm}
>{\centering\arraybackslash}m{0.8cm}
>{\centering\arraybackslash}m{3.2cm}
}
\toprule
Dataset    & Country         & \# Subject & Age      & Gender             & \# Activity & Device   & Location \\ 
\midrule
UCI      & Italy           & 30         & 19-48    & -                  & 6           & Custom   & Waist \\
\rowcolor{lightblue}
HHAR       & Denmark         & 9          & 25-30    & -                  & 6           & Custom   & wrist, waist \\
Shoaib     & The Netherlands & 10         & 25-30    & 10 male            & 7           & Custom   & wrist, pocket, waist, upperarm \\
\rowcolor{lightblue}
Motion     & England         & 24         & 18-46    & 14 male, 10 female & 6           & Custom   & pocket \\
DSADS      & Turkey          & 8          & 20-30    & 4 male, \quad 4 female   & 18          & Research & chest, wrist, knee \\
\rowcolor{lightblue}
USC-HAD    & United States   & 14         & 21-49    & -                  & 12          & Research & pocket \\
KU-HAR     & Bangladesh      & 90         & 18-34   & 75 male, 15 female & 18          & Custom   & waist \\
\rowcolor{lightblue}
PAMAP2     & Germany         & 9          & 23-31    & 8 male, 1 female  & 18          & Research & chest, ankle, wrist \\
TNDA-HAR   & China           & 50         & -        & -                  & 8           & Research & wrist, thigh, ankle, upperarm, back \\
\rowcolor{lightblue}
Mhealth    & Spain           & 10         & -        & -                  & 12          & Research & chest, wrist, ankle \\
WISDM      & United States   & 51         & -        & -                  & 18          & Custom   & wrist, pocket \\
\rowcolor{lightblue}
RealWorld  & Germany         & 15         & 19-44    & 8 male, 7 female & 8           & Custom   & chest, wrist, thigh, shin, waist, head, upperarm \\
HARSense   & India           & 12         & Above 23 & -                  & 6           & Custom   & - \\
\rowcolor{lightblue}
UT-Complex & The Netherlands & 10         & 23-35    & 10 male            & 13          & Custom   & wrist, pocket \\
\bottomrule
\end{tabular}
}
\label{tab:datasets}
\end{table}

%% file: tables/activities.tex
\begin{table}[H]
\centering
\footnotesize
\caption{Overview of 62 activities on the curated benchmark.}
\begin{tabular}{
|c|c||c|c||c|c|
}
\hline
\textbf{ID} & \textbf{Activity Name}
& \textbf{ID} & \textbf{Activity Name}
& \textbf{ID} & \textbf{Activity Name} \\
\hline
0 & sitting & 1 & standing & 2 & lying \\
\hline
3 & upstairs & 4 & downstairs & 5 & walking \\
\hline
6 & running & 7 & jumping & 8 & lying on right side \\
\hline
9 & talk sit & 10 & talk stand & 11 & stand sit \\
\hline
12 & lay stand & 13 & pick & 14 & walk backward \\
\hline
15 & walk circle & 16 & waist bends forward & 17 & frontal elevation of arms \\
\hline
18 & knees bending crouching & 19 & cycling & 20 & watching tv \\
\hline
21 & computer work & 22 & car driving & 23 & vacuum cleaning \\
\hline
24 & ironing & 25 & folding laundry & 26 & house cleaning \\
\hline
27 & sit to stand & 28 & sit to lie & 29 & lie to sit \\
\hline
30 & stand to lie & 31 & lie to stand & 32 & type \\
\hline
33 & write & 34 & coffee & 35 & talk \\
\hline
36 & smoke & 37 & eat & 38 & brushing teeth \\
\hline
39 & eating soup & 40 & eating chips & 41 & eating pasta \\
\hline
42 & drinking from cup & 43 & eating sandwich & 44 & clapping \\
\hline
101 & standing still in elevator & 102 & moving in elevator & 103 & elevator up \\
\hline
104 & elevator down & 201 & walking on flat treadmill & 202 & walking on inclined treadmill \\
\hline
203 & exercising on stepper & 204 & exercising on cross trainer & 205 & cycling on exercise bike \\
\hline
206 & rowing & 207 & playing basketball & 208 & push up \\
\hline
209 & sit up & 210 & table tennis & 211 & nordic walking \\
\hline
212 & playing soccer & 213 & tennis ball &  &  \\
\hline
\end{tabular}
\label{tab:activities}
\end{table}

%% file: tables/split.tex
\begin{table}[t]
\centering
\footnotesize
\caption{Split details of dataset-level five-fold cross-validation.}
\begin{tabular}{c|c}
\toprule
Split  & Datasets                    \\ \midrule
Fold 1 & DSADS, HHAR, HARSense       \\
Fold 2 & RealWorld, WISDM, Motion    \\
Fold 3 & PAMAP2, MHEALTH, UT-Complex \\
Fold 4 & KU-HAR, Shoaib, UCI         \\
Fold 5 & USC-HAD, TNDA-HAR           \\ \bottomrule
\end{tabular}
\label{tab:split}
\end{table}

%% file: tables/acc.tex
\begin{table}[H]
\centering
\footnotesize
\caption{Overall comparison results. Mean ± standard deviation of Accuracy (\%) from dataset-level five-fold cross-validation are reported. 
\textbf{Bold} indicates the best result within each row, and \underline{underline} indicates the best result within each column.}
\scalebox{0.68}{
\begin{tabular}{@{}c|c|c|c|cc|ccc|cc@{}}
\toprule

\multirow{2}{*}{Encoder} & \multirow{2}{*}{Classifier} & \multirow{2}{*}{Modality} & \multicolumn{1}{c|}{Recognition} & \multicolumn{2}{c|}{Reconstruction} & \multicolumn{3}{c|}{Contrastive} & \multicolumn{2}{c}{Hybrid} \\ \cmidrule(l){4-11}
& & & BioBankSSL & LIMU-BERT & CRT & TS-TCC & TS2Vec & FOCAL & SimMTM & CrossHAR \\ \midrule \midrule

\multirow{8}{*}{Transformer} 
& \multirow{2}{*}{CNN} 
& Acc  & 57.05 ± 5.31 & 59.07 ± 3.56 & \textbf{60.08 ± 4.50} & 31.79 ± 10.03 & 55.73 ± 4.93 & 57.76 ± 4.15 & 60.07 ± 4.36 & 59.38 ± 2.40 \\
& & Acc+Gyro & 50.91 ± 8.58 & 56.54 ± 5.52 & \textbf{58.56 ± 5.15} & 24.16 ± 6.16 & 50.37 ± 7.78 & 52.48 ± 7.22 & 55.80 ± 4.20 & 55.83 ± 5.30 \\

\cmidrule(l){2-11}
& \multirow{2}{*}{MLP}
& Acc  & \textbf{57.71 ± 5.17} & 51.50 ± 5.17 & 56.69 ± 6.13 & 23.28 ± 2.65 & 31.43 ± 3.99 & 56.58 ± 4.20 & 56.81 ± 4.97 & 52.48 ± 3.75 \\
& & Acc+Gyro & 52.62 ± 8.61 & 48.61 ± 9.65 & 42.18 ± 19.25 & 24.96 ± 5.41 & 30.55 ± 9.09 & 53.95 ± 9.19 & \textbf{55.04 ± 5.22} & 50.71 ± 6.03 \\

\cmidrule(l){2-11}
& \multirow{2}{*}{GRU}
& Acc  & 57.54 ± 4.95 & 58.58 ± 2.77 & \textbf{59.81 ± 4.38} & 22.90 ± 2.33 & 53.62 ± 3.17 & 55.73 ± 3.54 & 56.88 ± 5.11 & 58.82 ± 3.97 \\
& & Acc+Gyro & 52.13 ± 8.59 & 55.20 ± 7.62 & \textbf{59.42 ± 5.23} & 24.31 ± 5.83 & 49.48 ± 8.50 & 51.99 ± 7.87 & 52.99 ± 5.01 & 55.77 ± 5.37 \\

\cmidrule(l){2-11}
& \multirow{2}{*}{Transformer}
& Acc  & 57.09 ± 5.23 & 56.85 ± 3.38 & \textbf{60.86 ± 4.20} & 23.33 ± 2.18 & 50.73 ± 2.52 & 55.04 ± 4.38 & 57.14 ± 4.58 & 58.66 ± 3.11 \\
& & Acc+Gyro & 51.25 ± 8.01 & 54.35 ± 6.27 & \textbf{60.03 ± 4.91} & 25.16 ± 5.70 & 46.77 ± 10.55 & 51.66 ± 8.14 & 53.93 ± 5.52 & 55.87 ± 4.39 \\

\midrule

\multirow{8}{*}{ResNet}
& \multirow{2}{*}{CNN}
& Acc  & \textbf{\underline{63.19 ± 5.41}} & \underline{60.57 ± 4.04} & 56.28 ± 4.47 & 48.03 ± 3.38 & \underline{59.82 ± 3.69} & 62.99 ± 3.71 & 59.54 ± 4.09 & \underline{59.91 ± 4.39} \\
& & Acc+Gyro & 55.42 ± 6.11 & 56.35 ± 7.12 & 59.81 ± 4.85 & 46.41 ± 7.74 & 55.93 ± 5.36 & \textbf{61.71 ± 3.50} & 58.78 ± 5.08 & 55.74 ± 5.79 \\

\cmidrule(l){2-11}
& \multirow{2}{*}{MLP}
& Acc  & 59.23 ± 4.97 & 51.81 ± 3.10 & 42.02 ± 9.92 & 45.11 ± 2.78 & 54.29 ± 3.16 & \textbf{62.77 ± 4.00} & 58.42 ± 5.12 & 55.49 ± 5.80 \\
& & Acc+Gyro & 52.81 ± 8.84 & 47.70 ± 6.41 & 54.77 ± 6.60 & 37.01 ± 10.60 & 52.62 ± 8.37 & \textbf{63.00 ± 5.39} & 56.92 ± 5.04 & 51.65 ± 5.91 \\

\cmidrule(l){2-11}
& \multirow{2}{*}{GRU}
& Acc  & 59.58 ± 5.38 & 58.64 ± 3.69 & 56.03 ± 5.14 & 45.74 ± 2.64 & 58.41 ± 4.05 & \textbf{61.12 ± 3.71} & \underline{60.37 ± 4.33} & 58.87 ± 4.71 \\
& & Acc+Gyro & 54.93 ± 6.29 & 54.34 ± 6.73 & 59.91 ± 5.26 & 36.95 ± 11.61 & 53.26 ± 6.55 & \textbf{61.21 ± 3.44} & 54.61 ± 4.57 & 54.40 ± 8.49 \\

\cmidrule(l){2-11}
& \multirow{2}{*}{Transformer}
& Acc  & \textbf{62.45 ± 6.44} & 59.17 ± 3.99 & 57.64 ± 4.19 & 43.92 ± 2.14 & 57.93 ± 4.26 & 61.93 ± 3.85 & 59.65 ± 4.01 & 57.69 ± 3.68 \\
& & Acc+Gyro & 53.83 ± 6.26 & 54.41 ± 4.89 & \textbf{\underline{61.68 ± 4.71}} & 35.05 ± 9.32 & 53.26 ± 6.00 & 60.85 ± 2.56 & 53.24 ± 3.93 & 54.40 ± 6.84 \\

\midrule

\multirow{8}{*}{CNN}
& \multirow{2}{*}{CNN}
& Acc  & 57.82 ± 4.95 & 60.31 ± 3.73 & 58.40 ± 5.40 & \underline{50.93 ± 5.23} & 57.78 ± 2.99 & \textbf{61.03 ± 1.26} & 59.79 ± 4.98 & 58.92 ± 3.97 \\
& & Acc+Gyro & 51.73 ± 6.39 & 55.69 ± 5.37 & 59.04 ± 5.71 & 47.48 ± 5.27 & 54.89 ± 4.16 & \textbf{61.12 ± 3.82} & 56.51 ± 4.46 & 56.27 ± 4.53 \\

\cmidrule(l){2-11}
& \multirow{2}{*}{MLP}
& Acc  & 53.68 ± 5.46 & 51.77 ± 3.09 & 53.85 ± 6.99 & 48.74 ± 4.19 & 51.66 ± 4.07 & \textbf{61.85 ± 2.14} & 59.22 ± 4.66 & 55.52 ± 5.12 \\
& & Acc+Gyro & 46.95 ± 6.34 & 49.99 ± 6.31 & 54.32 ± 5.58 & 47.98 ± 5.44 & 48.10 ± 6.20 & \textbf{\underline{63.26 ± 4.37}} & 56.12 ± 4.60 & 54.79 ± 6.60 \\

\cmidrule(l){2-11}
& \multirow{2}{*}{GRU}
& Acc  & 57.76 ± 5.23 & 59.68 ± 4.92 & 58.64 ± 5.06 & 49.96 ± 5.10 & 56.98 ± 3.80 & \textbf{61.49 ± 2.25} & 59.30 ± 4.37 & 59.54 ± 5.07 \\
& & Acc+Gyro & 50.83 ± 7.47 & 52.82 ± 4.75 & 59.06 ± 6.02 & 47.34 ± 5.72 & 52.61 ± 4.43 & \textbf{62.43 ± 5.27} & 54.99 ± 4.87 & 55.99 ± 5.15 \\

\cmidrule(l){2-11}
& \multirow{2}{*}{Transformer}
& Acc  & 58.22 ± 3.97 & 59.20 ± 3.40 & 59.81 ± 4.42 & 49.07 ± 4.37 & 56.54 ± 3.21 & \textbf{61.69 ± 1.82} & 60.14 ± 5.38 & 58.47 ± 3.66 \\
& & Acc+Gyro & 51.92 ± 6.18 & 54.65 ± 4.64 & 60.47 ± 5.12 & 45.27 ± 4.97 & 51.93 ± 4.58 & \textbf{62.27 ± 4.33} & 56.08 ± 4.83 & 55.54 ± 5.13 \\

\bottomrule
\end{tabular}
}
\label{tab:overall_acc}
\end{table}

%% file: tables/cross-subject.tex
\begin{table}[h]
\centering
\footnotesize
\caption{Results of cross-subject evaluation. 
Mean ± standard deviation of accuracy and F1-score (\%) from subject-level five-fold cross-validation are reported.}
\label{tab:modality_results}
\begin{tabular}{c|c|c|c|c|c}
\toprule
Modality & Metric & BioBankSSL & CRT & FOCAL & SimMTM \\
\midrule
\multirow{2}{*}{Acc + Gyro}
& Accuracy & $72.18 \pm 1.39$ & $70.93 \pm 1.58$ & $71.57 \pm 1.83$ & $73.58 \pm 1.82$ \\
& F1       & $73.40 \pm 1.49$ & $71.80 \pm 1.75$ & $72.43 \pm 1.91$ & $74.83 \pm 2.13$ \\
\midrule
\multirow{2}{*}{Acc-only}
& Accuracy & $71.94 \pm 2.05$ & $63.98 \pm 3.67$ & $70.47 \pm 2.44$ & $71.90 \pm 2.20$ \\
& F1       & $73.55 \pm 1.74$ & $64.16 \pm 3.40$ & $70.70 \pm 2.34$ & $72.66 \pm 2.53$ \\
\bottomrule
\end{tabular}
\label{tab:subject}
\end{table}

%% file: tables/cross-device.tex
\begin{table}[h]
\centering
\footnotesize
\caption{Results of cross-device evaluation. 
Accuracy and F1-score (\%) of each method are reported.}
\label{tab:cross_domain}
\begin{tabular}{c|c|c|cc|cc|cc|cc}
\toprule
\multirow{2}{*}{Modality} & \multirow{2}{*}{Source} & \multirow{2}{*}{Target} 
& \multicolumn{2}{c|}{BioBankSSL} 
& \multicolumn{2}{c|}{CRT} 
& \multicolumn{2}{c|}{FOCAL} 
& \multicolumn{2}{c}{SimMTM} \\
\cmidrule(lr){4-5}
\cmidrule(lr){6-7}
\cmidrule(lr){8-9}
\cmidrule(l){10-11}
& & 
& Acc & F1 
& Acc & F1 
& Acc & F1 
& Acc & F1 \\
\midrule

\multirow{2}{*}{Acc + Gyro}
& Research & Custom 
& 47.00 & 45.00 
& 55.26 & 53.94 
& 53.47 & 51.12 
& 45.77 & 43.22 \\

& Custom & Research 
& 56.77 & 57.44 
& 59.76 & 59.51 
& 59.52 & 58.75 
& 52.01 & 49.48 \\

\midrule

\multirow{2}{*}{Acc-only}
& Research & Custom 
& 54.11 & 52.14 
& 52.76 & 51.10 
& 54.39 & 51.64 
& 49.68 & 48.15 \\

& Custom & Research 
& 57.53 & 58.52 
& 56.80 & 57.03 
& 52.97 & 52.40 
& 55.52 & 55.54 \\

\bottomrule
\end{tabular}
\label{tab:device}
\end{table}

%% file: tables/cross-location.tex
\begin{table}[h]
\centering
\footnotesize
\caption{Results of cross-location evaluation. 
Accuracy and F1-score (\%) of each method are reported.}
\label{tab:position}
\begin{tabular}{c|cc|cc|cc|cc|cc}
\toprule
\multirow{2}{*}{Modality} & \multirow{2}{*}{Source} & \multirow{2}{*}{Target}
& \multicolumn{2}{c|}{BioBankSSL}
& \multicolumn{2}{c|}{CRT}
& \multicolumn{2}{c|}{FOCAL}
& \multicolumn{2}{c}{SimMTM} \\
\cmidrule(lr){4-5}
\cmidrule(lr){6-7}
\cmidrule(lr){8-9}
\cmidrule(l){10-11}
& & 
& Acc & F1
& Acc & F1
& Acc & F1
& Acc & F1 \\
\midrule

\multirow{6}{*}{Acc + Gyro}
& Trunk & Upper limb 
& 49.25 & 50.29 
& 53.73 & 54.83 
& 53.35 & 53.22 
& 51.52 & 51.57 \\

& Trunk & Lower limb 
& 37.84 & 38.78 
& 51.68 & 51.51 
& 44.95 & 44.62 
& 37.72 & 36.91 \\

& Upper limb & Trunk 
& 47.10 & 46.49 
& 51.51 & 52.40 
& 51.13 & 50.27 
& 43.24 & 41.74 \\

& Upper limb & Lower limb 
& 39.71 & 38.51 
& 53.73 & 52.81 
& 41.78 & 40.07 
& 39.82 & 38.70 \\

& Lower limb & Trunk 
& 45.90 & 43.88 
& 52.17 & 53.21 
& 49.73 & 49.71 
& 46.26 & 44.42 \\

& Lower limb & Upper limb 
& 46.61 & 43.19 
& 52.37 & 50.72 
& 52.63 & 50.80 
& 49.49 & 46.08 \\

\midrule

\multirow{6}{*}{Acc-only}
& Trunk & Upper limb 
& 48.38 & 49.20 
& 52.95 & 54.66 
& 48.22 & 48.41 
& 46.16 & 45.61 \\

& Trunk & Lower limb 
& 39.91 & 38.33 
& 45.07 & 43.76 
& 38.24 & 35.86 
& 36.97 & 35.26 \\

& Upper limb & Trunk 
& 47.78 & 47.40 
& 50.30 & 51.61 
& 51.13 & 50.26 
& 49.23 & 49.02 \\

& Upper limb & Lower limb 
& 41.16 & 40.10 
& 47.71 & 48.20 
& 41.78 & 40.07 
& 43.69 & 42.00 \\

& Lower limb & Trunk 
& 43.86 & 40.93 
& 48.26 & 47.25 
& 48.36 & 47.19 
& 43.90 & 42.01 \\

& Lower limb & Upper limb 
& 43.77 & 40.38 
& 47.96 & 47.43 
& 48.07 & 45.82 
& 46.31 & 42.16 \\

\bottomrule
\end{tabular}
\label{tab:location}
\end{table}